\documentclass{article}

\usepackage{microtype}
\usepackage{graphicx}
\usepackage{subcaption}
\usepackage{booktabs}
\usepackage{hyperref}

\usepackage[accepted]{icml2026}
\providecommand{\RETURN}{\STATE \textbf{return}~}

\usepackage{amsmath}
\usepackage{amssymb}
\usepackage{mathtools}
\usepackage{enumitem}
\usepackage{xcolor}
\usepackage{xspace}
\usepackage{algorithm}
\usepackage{algorithmic}
\usepackage{arydshln}
\usepackage{amsthm}
\newtheorem{proposition}{Proposition}

\definecolor{cgreen}{HTML}{15803D}
\definecolor{cblue}{HTML}{1D4ED8}
\definecolor{cyellow}{HTML}{92400E}
\definecolor{cpink}{HTML}{9D174D}

\newcommand{\seva}{\textsc{Seva}\xspace}
\newcommand{\realnumber}[1]{\textbf{#1}}

\icmltitlerunning{SEVA: Self-Evolving Verification Agent with Process Reward}

\begin{document}

\twocolumn[
  \icmltitle{SEVA: Self-Evolving Verification Agent with\\Process Reward for Fact Attribution}

  \begin{icmlauthorlist}
  \icmlauthor{Aojie Yuan}{usc}
  \icmlauthor{Yi Nian}{usc}
  \icmlauthor{Haiyue Zhang}{usc}
  \icmlauthor{Zijian Su}{umich}
  \icmlauthor{Yue Zhao}{usc}
  \end{icmlauthorlist}

  \icmlaffiliation{usc}{University of Southern California, Los Angeles, USA}
  \icmlaffiliation{umich}{University of Michigan, Ann Arbor, USA}

  \icmlcorrespondingauthor{Aojie Yuan}{aojieyua@usc.edu}
  \icmlcorrespondingauthor{Yi Nian}{yinian@usc.edu}
  \icmlcorrespondingauthor{Haiyue Zhang}{haiyuez@usc.edu}
  \icmlcorrespondingauthor{Zijian Su}{simoon@umich.edu}
  \icmlcorrespondingauthor{Yue Zhao}{yue.z@usc.edu}

  \icmlkeywords{fact verification, hallucination detection, process reward, GRPO, structured output, self-evolution, verification agent}

  \vskip 0.3in
]

% Override the AIWILD-style default notice with the AI4GOOD venue
\makeatletter
\renewcommand{\ICML@appearing}{%
  \textit{Accepted at the ICML 2026 Workshop on Trustworthy AI for Good (AI4GOOD), Seoul, South Korea.}
  Copyright 2026 by the author(s).%
  \quad Code: \url{https://github.com/Justin0504/Verifiable_agent}%
}
\makeatother

\printAffiliationsAndNotice{}

% ==============================================================
\begin{abstract}
% ==============================================================
Hallucination is the reliability bottleneck for LLM-based agents, and fact attribution verifiers are the last line of defense --- yet today's verifiers emit only opaque binary labels, leaving agents unable to self-correct and operators unable to audit.
We present \seva, a structured verification agent that emits evidence alignments, step-by-step reasoning chains, calibrated confidence, and a six-category error diagnosis with actionable fixes.
Training such an agent with RL is non-trivial: standard binary reward on multi-component output triggers \emph{advantage collapse} --- within-group reward variance vanishes and the GRPO gradient disappears.
We resolve this with a process reward that decomposes verification quality into five independent components weighted $70/30$ toward process signals, restoring the gradient and inducing an implicit curriculum --- the agent first masters verification \emph{behavior} (alignment $0.917{\to}0.997$, format $72\%{\to}100\%$), then \emph{outcomes} (F1 $64.9{\to}69.0$).
Structured output further enables a Verify$\to$Reflect$\to$Probe$\to$Refine self-evolution loop, which over four rounds on a 7B model surfaces an unexpected structural finding: each round produces a \emph{benchmark-specialist}, not a generalist ($+15$\,pp on HaluEval, $-10$ to $-14$\,pp on TruthfulQA in the same model, persistent at $4\times$ data).
On ClearFacts, \seva-3B matches GPT-4o-mini (69.0 vs.\ 69.8 F1) while producing substantially richer, auditable output --- confirming a principle that should generalize: for any RL task with multi-component generation, \emph{reward granularity must match output granularity}.
\end{abstract}

% ==============================================================
\section{Introduction}
\label{sec:intro}
% ==============================================================

Despite rapid progress in LLM capabilities, hallucination remains a fundamental barrier to deploying agents in high-stakes domains such as finance, law, and healthcare~\citep{min2023factscore}.
Fact attribution verifiers --- models that judge whether each claim in an agent's output is supported by its source documents --- have emerged as a critical safety layer~\citep{gao2023rarr,tian2024finetuning}.
Systems like MiniCheck~\citep{tang2024minicheck} and ClearCheck~\citep{seo2025vtv} achieve strong accuracy on this task, but they share a fundamental limitation: they produce only a binary label.

This opacity creates two problems for agents in the wild.
First, when a verifier flags a claim, the agent has no basis for \emph{self-correction} --- it knows something is wrong, but not whether a percentage was inflated, an entity swapped, or a qualifier fabricated.
Second, no human operator can meaningfully \emph{audit} the decision, because the reasoning behind the label is invisible.
In safety-critical deployments, an uninterpretable verifier undermines the very trust it is meant to provide.

We introduce \seva (\textbf{S}elf-\textbf{E}volving \textbf{V}erification \textbf{A}gent), which addresses both problems by producing \emph{structured} verification output: evidence alignment spans that ground every judgment in specific text, reasoning chains that trace the logic step by step, and error diagnoses from a six-category taxonomy with actionable fix suggestions.
This structured output serves a dual purpose: it makes verification auditable for deployment, and it provides a diagnostic interface for training.

How should such an agent be trained?
SFT on teacher-annotated data provides a reasonable starting point, but reinforcement learning --- which has driven substantial gains for mathematical reasoning~\citep{shao2024deepseekmath,zha2025tango} and hallucination reduction~\citep{li2026march} --- does not straightforwardly transfer.
Applying GRPO~\citep{shao2024deepseekmath} with binary reward (1 if the label matches, 0 otherwise) to our structured verifier, we find that training stalls entirely: the policy makes no progress beyond SFT across all 350 steps.
The culprit is that binary reward compresses all verification quality into a single bit.
A response with correct reasoning but the wrong label receives the same score --- zero --- as one that produces unparseable garbage.
In a GRPO group of $G{=}8$ responses, most therefore score 0, advantage spread contracts to $\pm 0.05$, and the gradient vanishes.

Our response is a \textbf{process reward function} $R: \mathcal{V} \times \mathcal{Y} \to [-0.10, 1.28]$ mapping each structured response $\mathbf{v}$ (against gold $y^*$) to a continuous score over five independent components plus a calibration term:
\begin{equation}
\small
R \,=\, \underbrace{w_f R_f \!+\! w_a R_a \!+\! w_c R_c}_{\text{process }(70\%)} \!+\! \underbrace{w_l R_l \!+\! w_d R_d}_{\text{outcome }(30\%)} \!+\! R_{\text{cal}}
\label{eq:reward}
\end{equation}
with weights $w_f{=}0.10$, $w_a{=}w_c{=}0.30$, $w_l{=}w_d{=}0.15$, and an asymmetric calibration $R_{\text{cal}} = +\hat{\gamma}{\cdot}0.15$ if $\hat{y}{=}y^*$ else $-\hat{\gamma}{\cdot}0.10$, which rewards calibrated correctness more than it penalizes calibrated error.
Each $R_x \in [0, 1]$ is computed independently from a different region of $\mathbf{v}$ (App.~\ref{app:reward}), so the components are weakly correlated; this independence is what creates the smooth four-level reward landscape (Fig.~\ref{fig:landscape}, Tab.~\ref{tab:reward_landscape}) that resolves the collapse.
A response with sound reasoning but the wrong label scores $\sim 0.63$ rather than $0.0$, restoring $\sigma > 0$ in Eq.~\ref{eq:advantage}.

The results confirm that process reward unlocks what binary reward cannot.
GRPO lifts alignment quality to 0.997, format compliance to 100\%, and F1 to 69.0 (+4.1 over SFT).
An implicit curriculum emerges in the training dynamics: the agent masters verification \emph{behavior} within 150 steps, then spends the remaining 200 steps refining verification \emph{outcomes} --- without any explicit scheduling.

Structured output confers a further advantage: it makes the agent's failures \emph{transparent}.
When \seva misclassifies a claim, its evidence alignments and error diagnoses pinpoint which error category was missed and where grounding broke down.
We channel this diagnostic signal into a \textbf{Verify}$\to$\textbf{Reflect}$\to$\textbf{Probe}$\to$\textbf{Refine} self-evolution loop that generates targeted adversarial data for the agent's weakest error categories, and iterate it four times on the 7B model.
A surprising empirical finding emerges: each round yields a \emph{benchmark-specialist} rather than a strictly better generalist, and the asymmetric trade-off persists at $4\times$ training-data scale ($7{,}787$ samples in Round~4) --- confirming that the effect is data-distribution-induced rather than overfitting.
This finding is itself only visible because the structured output exposes per-category error dynamics that aggregate accuracy would hide, and it sits uneasily with the monotonic-improvement assumption implicit in Self-Refine / STaR-style self-evolution literature.

Our contributions are threefold.
\begin{enumerate}[nosep,leftmargin=*]
  \item \textbf{\seva: a structured verifier that's also auditable.} A 3B agent whose output --- alignments, reasoning chain, calibrated confidence, six-category error diagnosis with fixes --- matches GPT-4o-mini's accuracy ($69.0$ vs.\ $69.8$ F1) while producing the substrate every downstream operator needs (\S\ref{sec:format},~\S\ref{sec:experiments}).
  \item \textbf{A process reward that turns RL on structured output from impossible to possible.} We formalize \emph{advantage collapse} (Prop.~\ref{prop:collapse}) as the failure mode of binary reward on multi-component generation, then resolve it with a five-component decomposition (Prop.~\ref{prop:variance}); the resulting reward landscape yields an \emph{implicit curriculum} --- behavior before outcomes --- without any explicit scheduling (\S\ref{sec:reward},~\S\ref{sec:implicit}).
  \item \textbf{A self-evolution loop that reveals a structural property of iterative refinement.} Verify$\to$Reflect$\to$Probe$\to$Refine, iterated four rounds on a 7B model, surfaces a finding that contradicts the monotone-improvement assumption of Self-Refine / STaR: each round produces a \emph{benchmark-specialist}, not a generalist --- $+15$\,pp HaluEval, $-10$ to $-14$\,pp TruthfulQA in the \emph{same} model, robust at $4{\times}$ training data, visible only because structured output exposes per-category dynamics (\S\ref{sec:evolution}).
\end{enumerate}

% ==============================================================
\section{Method}
\label{sec:method}
% ==============================================================

\begin{figure*}[t]
\centering
\includegraphics[width=\textwidth]{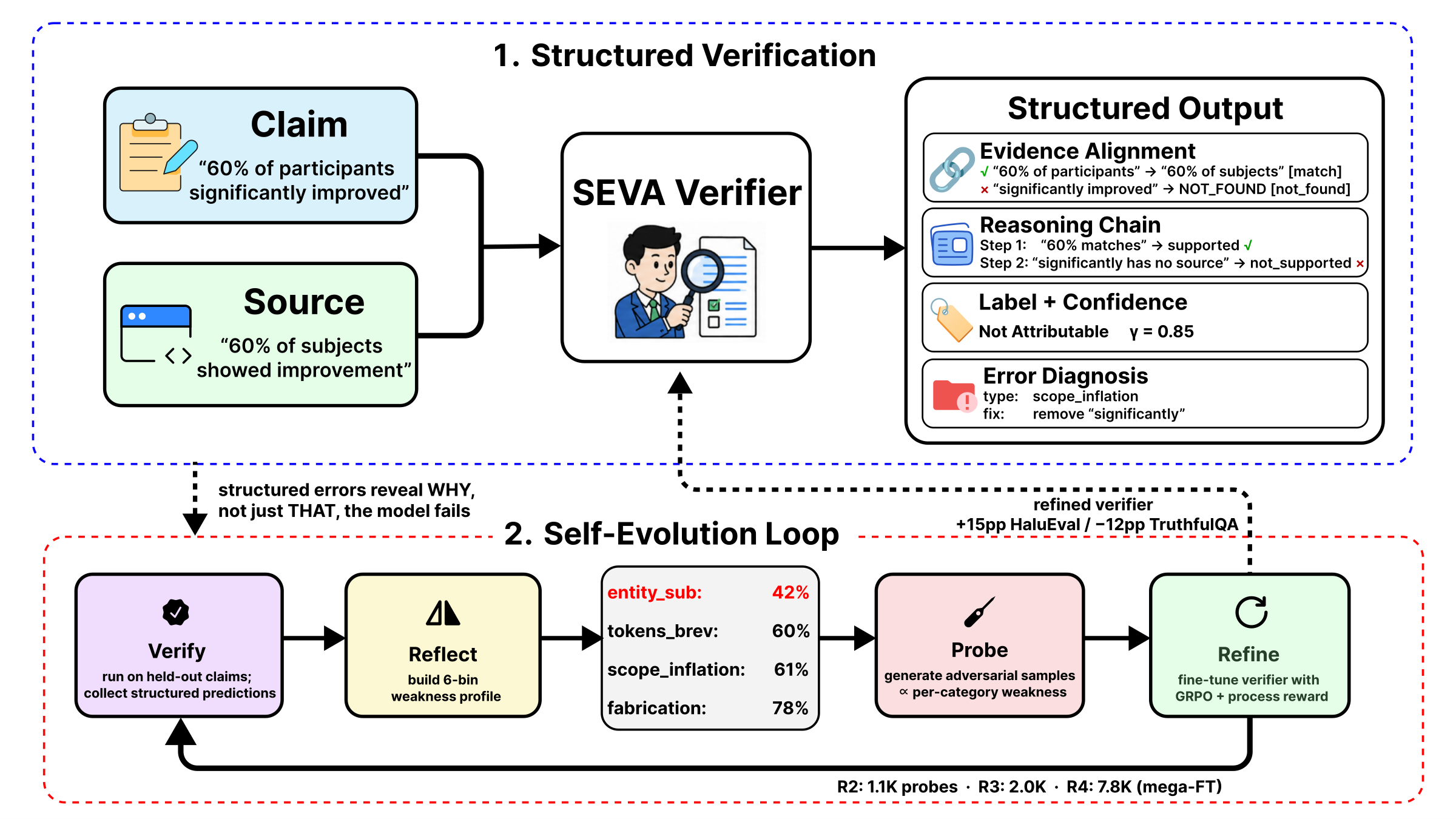}
\caption{\textbf{\seva overview.} \textit{Top:} Given a claim-source pair, the verifier produces structured output --- evidence alignments, reasoning chains, calibrated confidence, and error diagnosis. \textit{Bottom:} Self-evolution loop. Structured errors reveal \emph{why} the model fails (not just \emph{that} it fails), enabling targeted adversarial data generation focused on the weakest error types.}
\label{fig:overview}
\end{figure*}

\subsection{Problem Formulation and Overview}
\label{sec:overview}

Given a claim $c$ and source document $d$, a fact attribution verifier produces a judgment about whether $c$ is supported by $d$.
Existing verifiers output a single binary label~\citep{tang2024minicheck,seo2025vtv}.
We instead require the agent to produce structured output $\mathbf{v} = (A, C, y, \gamma, e, s)$ that makes verification auditable and failures diagnosable (Figure~\ref{fig:overview}).

Building such an agent via SFT is straightforward, but pushing it further with RL is not.
We show that GRPO with binary reward fails entirely on this output format (\S\ref{sec:negative}), and design a process reward that resolves this failure (\S\ref{sec:reward}).
The structured output further enables a self-evolution loop for iterative improvement (\S\ref{sec:evolution}).

\subsection{Structured Verification Schema}
\label{sec:format}

The output $\mathbf{v}$ comprises four complementary components:

\textbf{Evidence alignment} $A$: a list of $(c_i, d_i, \text{status}_i)$ triples mapping claim spans to source spans, with status $\in$ \{match, mismatch, not\_found\}.
Each entry forces the agent to anchor its judgment in specific text rather than forming a holistic impression.

\textbf{Reasoning chain} $C$: step-by-step verification where each step examines a claim part against source evidence, producing a judgment $\in$ \{supported, not\_supported, partially\_supported\} and a natural language explanation.

\textbf{Label and confidence}: a binary label $y$ paired with calibrated confidence $\gamma \in [0,1]$.

\textbf{Error diagnosis}: when $y{=}$Not Attributable, an error type $e$ drawn from a six-category taxonomy (numerical exaggeration, negation flip, scope inflation, temporal shift, entity substitution, fabrication) together with a fix suggestion $s$.

This schema serves two purposes relevant to agents in the wild.
First, it makes verification \emph{auditable}: a human operator can inspect alignments and reasoning to judge whether the verdict is trustworthy.
Second, it makes failures \emph{diagnosable}: when the agent errs, the structured output pinpoints which evidence was mishandled, feeding the self-improvement loop in~\S\ref{sec:evolution}.

\subsection{From Binary Reward Failure to Process Reward}
\label{sec:negative}\label{sec:reward}

\paragraph{The failure of binary reward.}
Binary reward assigns 1.0 when the predicted label matches the gold label and 0.0 otherwise.
For structured output, this produces a degenerate training signal.
In a GRPO group of $G{=}8$ responses:
(1)~28\% fail JSON parsing --- the SFT model produces valid JSON only 72\% of the time, and all of these score 0;
(2)~among valid responses, $\sim$35\% predict the wrong label, also scoring 0;
(3)~in a typical group, 5--7 of 8 responses receive zero reward.
GRPO computes advantages relative to the group mean.
For a group of $G$ responses with rewards $\{r_1, \ldots, r_G\}$, the normalized advantage of response $i$ is:
\begin{equation}
\hat{A}_i = \frac{r_i - \mu}{\sigma + \epsilon}, \quad \mu = \frac{1}{G}\sum_{j} r_j, \quad \sigma = \text{std}(\{r_j\})
\label{eq:advantage}
\end{equation}
When binary reward produces $r_j \in \{0, 1\}$ with most $r_j = 0$, both $\mu$ and $\sigma$ are near zero, and $\hat{A}_i \approx 0$ for all $i$ --- the policy gradient $\nabla_\theta J \propto \sum_i \hat{A}_i \nabla_\theta \log \pi_\theta$ vanishes regardless of model parameters.

This failure is \emph{structural}, not incidental.
Increasing group size does not help --- the problem is near-uniform scores, not insufficient sampling.
And the failure is not specific to verification: any RL task whose output has multiple required components will exhibit advantage collapse under binary reward whenever the model cannot reliably produce all components simultaneously.
We formalize the mechanism below; a proof sketch is given in Appendix~\ref{app:theory}.

\begin{proposition}[Binary-Reward Advantage Collapse]
\label{prop:collapse}
Let $r_1, \ldots, r_G \in \{0, 1\}$ be i.i.d.\ Bernoulli rewards in a GRPO group of size $G$ with success probability $q = \Pr[r_j {=} 1]$. The expected within-group variance is
\begin{equation}
\mathbb{E}[\sigma^2] \,=\, \tfrac{G}{G-1}\, q\,(1-q),
\label{eq:bernoulli-var}
\end{equation}
and as $q \to 0^+$ or $q \to 1^-$, $\sigma \xrightarrow{a.s.} 0$, hence
\begin{equation}
\hat{A}_i \,=\, \frac{r_i - \mu}{\sigma + \epsilon} \,\xrightarrow{a.s.}\, 0 \qquad \forall\, i,
\label{eq:adv-collapse}
\end{equation}
and the expected policy gradient
\begin{equation}
\nabla_\theta J \,=\, \mathbb{E}\!\left[\sum_{i=1}^{G} \hat{A}_i \, \nabla_\theta \log \pi_\theta(\mathbf{v}_i)\right] \,\to\, 0
\label{eq:grad-vanish}
\end{equation}
\emph{regardless of} model parameters $\theta$ or group size $G$.
\end{proposition}

\begin{proposition}[Process-Reward Variance Lower Bound]
\label{prop:variance}
Let $R = \sum_{k=1}^{K} w_k R_k$ be the aggregate process reward with components $R_k \in [0,1]$ and positive weights $w_k > 0$. By the variance identity for linear combinations,
\begin{equation}
\sigma^2(R) \,=\, \sum_{k=1}^{K} w_k^2\, \sigma_k^2 \,+\, 2\sum_{k<\ell} w_k\, w_\ell \,\mathrm{Cov}(R_k, R_\ell).
\label{eq:variance-decomp}
\end{equation}
Unless the components are perfectly anti-correlated, the cross-terms cannot drive $\sigma^2(R)$ to zero unless every $\sigma_k {=} 0$. In particular, if any single component $k^*$ has $\sigma_{k^*}^2 > 0$ and is uncorrelated with the rest,
\begin{equation}
\sigma^2(R) \,\geq\, w_{k^*}^2\, \sigma_{k^*}^2 \,>\, 0,
\label{eq:variance-lower}
\end{equation}
and the GRPO gradient is non-vanishing.
\end{proposition}

\paragraph{Why these matter.}
Prop.~\ref{prop:collapse} pinpoints \emph{the} failure mode of binary reward on structured output, and Prop.~\ref{prop:variance} guarantees that process reward escapes it by construction.
Empirically, $q \approx 0.37$ at SFT-init gives $\mathbb{E}[\sigma^2] \lesssim 0.27$ (Eq.~\ref{eq:bernoulli-var}) and shrinks to $\sigma \approx 0.05$ by step $350$ (Tab.~\ref{tab:dynamics_full}) --- the gradient effectively dies under binary reward.
Under process reward, format errors at $\sim 28\%$ keep $\sigma_f > 0$ at every step we observe, so Eq.~\ref{eq:variance-lower} delivers $\sigma^2(R) > 0$ throughout training; the smooth four-level landscape of Tab.~\ref{tab:reward_landscape} is the geometric consequence.
The argument is task-agnostic: any RL setting whose output has $K$ required components inherits the same dichotomy, so a process-style decomposition is the structural fix wherever it applies.

\begin{figure*}[t]
\centering
\includegraphics[width=0.96\textwidth]{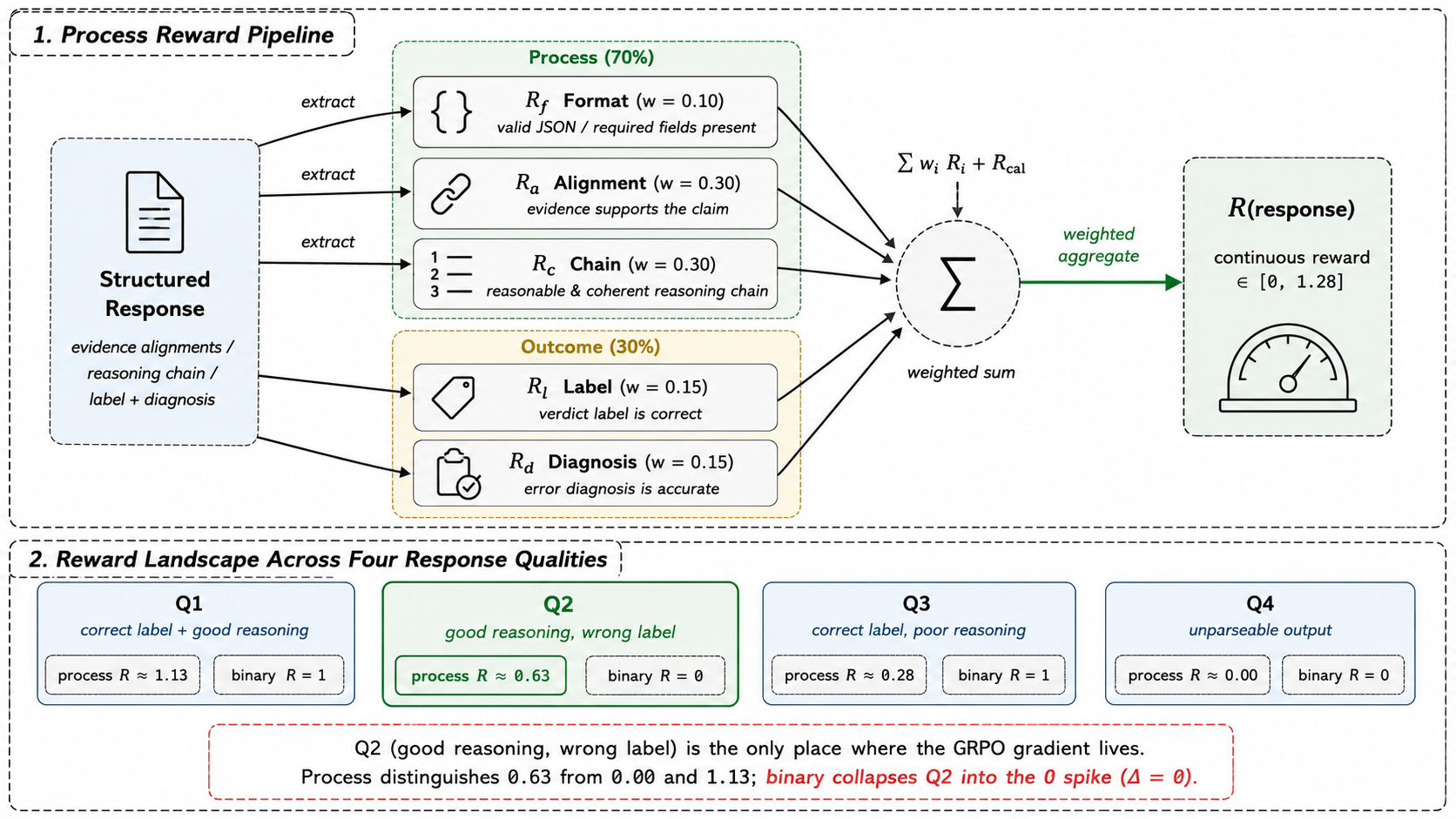}
\caption{\textbf{Process reward scoring.} Each structured output component (left) maps to an independently scored reward term (right). A response with correct reasoning but the wrong label scores 0.63 under process reward vs.\ 0.0 under binary --- this gap is what provides GRPO with meaningful gradients.}
\label{fig:reward}
\end{figure*}

The reward landscape (Tab.~\ref{tab:reward_landscape}) inverts the binary ranking: ``good reasoning, wrong label'' scores $0.63$ (vs.\ $0$) and ``correct label, poor reasoning'' only $0.28$ (vs.\ $1$) --- binary reward effectively pays for lucky guesses; process reward pays for genuine verification work.

\begin{table}[t]
\centering
\small
\caption{Reward landscape. Process reward produces four distinct quality levels where binary reward sees only two, enabling fine-grained advantage estimation within each GRPO group.}
\label{tab:reward_landscape}
\vspace{-4pt}
\begin{tabular}{lcc}
\toprule
\textbf{Response quality} & \textbf{Process} & \textbf{Binary} \\
\midrule
Correct label + good reasoning & $\sim$1.13 & 1.0 \\
Good reasoning, wrong label & $\sim$0.63 & 0.0 \\
Correct label, poor reasoning & $\sim$0.28 & 1.0 \\
Unparseable output & 0.0 & 0.0 \\
\bottomrule
\end{tabular}
\end{table}

\paragraph{Why 70/30, and how each component scores.}
The split forces the agent to do substantive verification \emph{before} the label becomes the easy lever: with outcome dominating, the model would learn to guess labels and produce incoherent reasoning around them.
Each $R_x$ scores independently --- $R_f$ on JSON validity, $R_a$ on per-span grounding, $R_c$ on per-step judgment and citation, $R_l$ on label match, $R_d$ on error type and fix; an asymmetric calibration term $R_{\text{cal}}\!\in\!\{+\gamma{\cdot}0.15, -\gamma{\cdot}0.10\}$ rewards confident correctness and penalizes overconfident error.
Algorithm~\ref{alg:reward} gives the full computation; per-component rubrics are in Appendix~\ref{app:reward}.

\begin{algorithm}[t]
\small
\caption{Process Reward Computation}
\label{alg:reward}
\begin{algorithmic}[1]
\REQUIRE Response $r$, gold label $y^*$
\ENSURE Reward $R \in [-0.10, 1.28]$
\STATE Parse $r$ as JSON $\to$ $\hat{v}$
\IF{parse fails}
  \RETURN $R = 0.0$
\ENDIF
\STATE $R_f \gets \text{ScoreFormat}(\hat{v})$ \COMMENT{0 / 0.2 / 0.5 / 1.0}
\STATE $R_a \gets \frac{1}{|A|}\sum_{a \in A} \text{ScoreAlign}(a)$
\STATE $R_c \gets \frac{1}{|C|}\sum_{s \in C} \text{ScoreStep}(s) + \text{LenBonus}$
\STATE $R_l \gets \mathbf{1}[\text{normalize}(\hat{y}) = y^*]$
\STATE $R_d \gets \text{ScoreDiagnosis}(\hat{e}, \hat{s}, y^*)$
\STATE $R \gets 0.10\, R_f + 0.30\, R_a + 0.30\, R_c + 0.15\, R_l + 0.15\, R_d$
\IF{$\hat{y} = y^*$}
  \STATE $R \gets R + \hat{\gamma} \times 0.15$ \COMMENT{calibration bonus}
\ELSE
  \STATE $R \gets R - \hat{\gamma} \times 0.10$ \COMMENT{overconfidence penalty}
\ENDIF
\RETURN $R$
\end{algorithmic}
\end{algorithm}

\subsection{Training Pipeline}
\label{sec:training}

\seva is trained in two phases.
\textbf{SFT:} GPT-4o-mini annotates $4{,}992$ ANLI examples with structured output ($92\%$ format-valid); Qwen2.5-3B-Instruct~\citep{qwen2025qwen25} is fine-tuned for $3$ epochs at lr $=2{\times}10^{-5}$.
\textbf{GRPO:} the SFT checkpoint seeds 5 epochs ($\sim$350 steps) of process-reward GRPO with $G{=}8$, $T{=}1.2$, $\beta{=}0.001$, lr $=2{\times}10^{-6}$, on veRL~\citep{sheng2024hybridflow} with FSDP on 2$\times$RTX 6000 Ada ($\sim$28 GPU-hours total).
The low GRPO lr and small $\beta$ jointly preserve the SFT-established format while leaving room for the policy to explore; we found this balance via a small sweep ($\beta{=}0.01$ over-regularized; $\beta{=}0$ admitted format-gaming).
We additionally train a 7B variant via two-stage SFT (binary NLI $\to$ structured) with LoRA-128~\citep{hu2022lora}; full hyperparameters in Appendix~\ref{app:impl}.

\subsection{Self-Evolution via Structured Diagnostics}
\label{sec:evolution}

Structured output exposes which aspect of verification failed when the agent misclassifies; we channel this signal through a \textbf{Verify}$\to$\textbf{Reflect}$\to$\textbf{Probe}$\to$\textbf{Refine} loop (Fig.~\ref{fig:overview}, bottom), borrowing the principle of \emph{functional separation} from MARCH~\citep{li2026march} but applying it across loop stages rather than agents.
Reflect aggregates error diagnoses into a 6-bin weakness profile; Probe allocates adversarial generation budget proportional to per-category weakness, giving weak bins $\sim$$3\times$ the budget of strong ones (e.g., entity\_sub at $42\%$ acc gets $\sim$$3\times$ fabrication's at $78\%$).

GRPO training itself constitutes Round~0: the process reward continuously assesses structural quality, and rollout sampling explores the agent's decision boundary.
We then iterate four additional rounds on the 7B Step150 seed:
\textbf{Round~1} injects extracted verification rules into the prompt (no parameter update);
\textbf{Round~2} performs LoRA SFT on $1{,}122$ adversarial probes;
\textbf{Round~3} performs full FT on $2{,}013$ mixed samples (adversarial $+$ replay);
\textbf{Round~4} (``mega-FT'') extends Round~3 with $7{,}787$ mixed samples to test whether more diverse adversarial data closes any remaining gap.
Pseudocode for the four-stage loop is in App.~\ref{app:se_full} (Algorithm~\ref{alg:sevolve}).

\begin{figure}[t]
\centering
\includegraphics[width=\columnwidth]{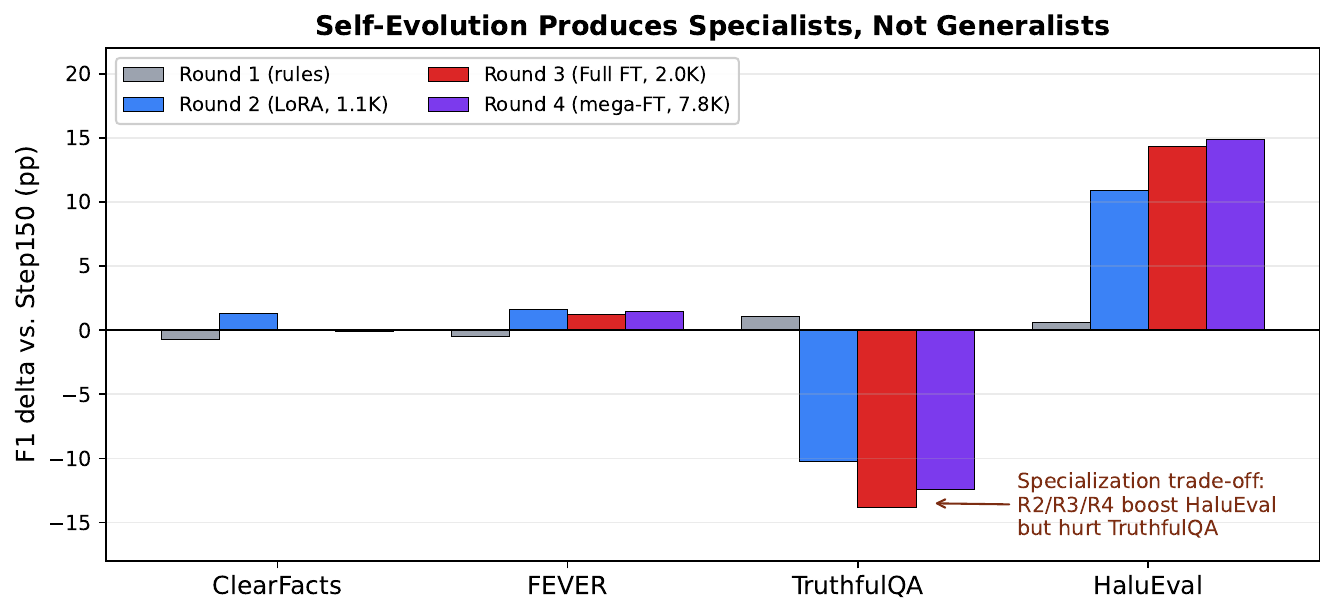}
\caption{\textbf{Self-evolution produces specialists, not generalists.} F1 deltas vs.\ the Step150 GRPO seed across four benchmarks and four refinement rounds. The asymmetric trade-off on TruthfulQA vs.\ HaluEval (Rounds~2--4) sharpens with each round and persists at $4\times$ training-data scale (Round~4, $7{,}787$ samples), confirming the effect is data-distribution-induced rather than overfitting. Absolute F1 numbers in Appendix~\ref{app:se_full}, Table~\ref{tab:se_full}.}
\label{fig:specialization}
\end{figure}

\paragraph{Specialists, not generalists.}
We expected four refinement rounds to yield a monotonically improving generalist; we observe a sharper \emph{specialization fingerprint} instead (Table~\ref{tab:specialization}).
Round~2 lifts HaluEval by $+10.9$\,pp but drops TruthfulQA by $-10.2$; Round~3 sharpens to $+14.3 / -13.8$; Round~4 holds at $+14.9 / -12.4$ despite $4\times$ more data.
Persistence at $4\times$ scale rules out trivial overfitting and identifies the effect as \emph{data-distribution-induced}: probes drawn from a ClearFacts-style weakness profile push the model toward those failure modes and away from TruthfulQA's distribution.
\emph{When probes come from a single source distribution, specialization is the dominant mode of iterative refinement.}

\begin{table}[h]
\centering
\footnotesize
\caption{Self-evolution produces \emph{specialists}, not generalists. Each row shows macro-F1 deltas vs.\ Step150 across four benchmarks. The opposite-sign gains on TruthfulQA vs.\ HaluEval (rows 2--4) persist at $4\times$ training-data scale (Round~4), confirming the effect is data-distribution-induced.}
\label{tab:specialization}
\setlength{\tabcolsep}{3pt}
\begin{tabular}{@{}lcccc@{}}
\toprule
& \textbf{ClearF.} & \textbf{FEVER} & \textbf{TrQA} & \textbf{HaluE.} \\
\midrule
Round 1 (rules)    & $-$0.7 & $-$0.5 & $+$1.1 & $+$0.6 \\
Round 2 (LoRA)     & $+$1.3 & $+$1.6 & \textbf{$-$10.2} & $+$10.9 \\
Round 3 (FT)       & 0.0    & $+$1.2 & \textbf{$-$13.8} & $+$14.3 \\
Round 4 (mega-FT)  & $-$0.1 & $+$1.5 & \textbf{$-$12.4} & $+$14.9 \\
\bottomrule
\end{tabular}
\end{table}

\paragraph{Does the loop actually work?}
Three properties make the data-volume reading hard to sustain: \emph{(i)} the trade-off is monotone in both directions ($+10.9{\to}+14.9$ HaluEval, $-10.2{\to}-12.4$ TruthfulQA --- a non-functional loop would sign-flip); \emph{(ii)} per-benchmark gains track Probe-stage budget allocation, not raw sample count; \emph{(iii)} Round~4 has $7\times$ Round~2's data but adds only $+4$\,pp on HaluEval, exhibiting saturation rather than the unbounded growth of data-volume overfitting.
The clean isolation ablation --- swapping the weakness-guided Probe for a same-budget random sampler --- is out of scope for this submission (\S\ref{sec:limitations}); the loop's response to its signal is consistent with a working mechanism (full analysis in App.~\ref{app:se_full}).
This specialization fingerprint is itself only visible because structured output exposes per-category dynamics; the same asymmetry would be invisible in aggregate accuracy, motivating downstream architectural responses (e.g., per-domain routing across rounds) we explore in follow-up work.

% ==============================================================
\section{Experiments}
\label{sec:experiments}
% ==============================================================

Having laid out the architecture (\S\ref{sec:format}), the reward (\S\ref{sec:reward}), and the self-evolution loop (\S\ref{sec:evolution}), we now stress-test \seva on four axes that any deployed verifier must clear: \textbf{accuracy} against established binary baselines, \textbf{generalization} across benchmarks with different failure modes, \textbf{structural reliability} of the produced output, and \textbf{training dynamics} under both reward designs.

\subsection{Setup}

We evaluate on ClearFacts~\citep{seo2025vtv} (1,590 samples; our primary metric), FEVER~\citep{thorne2018fever} (200), TruthfulQA~\citep{lin2022truthfulqa} (400), and HaluEval~\citep{li2023halueval} (200).
Together these cover claim-source attribution, encyclopedic verification, common misconceptions, and LLM-generated hallucinations --- four distinct distributions chosen to probe whether \seva's structural advantages survive across error types.

Baselines include binary verifiers reported by~\citet{seo2025vtv}: MiniCheck-7B (81.2 F1), ClearCheck-8B ($\sim$84 F1), and Llama-3.1-8B zero-shot (67.2 F1).
For structured comparison we evaluate GPT-4o-mini with zero-shot SEVA prompting and MiniCheck-Flan-T5-Large (770M).
We report macro F1, accuracy, and structural quality (alignment quality $R_a$, chain quality $R_c$, format compliance rate).

\subsection{Main Results}

\begin{table}[t]
\centering
\small
\caption{ClearFacts results. \seva models produce full structured output; binary baselines from~\citet{seo2025vtv}. The gap with MiniCheck reflects training data scale (5K structured vs.\ 57K binary) rather than a limitation of the approach.}
\label{tab:main}
\setlength{\tabcolsep}{3pt}
\begin{tabular}{llccc}
\toprule
\textbf{Model} & \textbf{Size} & \textbf{Output} & \textbf{Acc} & \textbf{F1} \\
\midrule
\multicolumn{5}{l}{\textit{Binary-label verifiers}} \\
Llama-3.1 (0-shot) & 8B & binary & -- & 67.2 \\
MiniCheck & 7B & binary & -- & 81.2 \\
ClearCheck & 8B & binary & -- & $\sim$84 \\
\midrule
\multicolumn{5}{l}{\textit{Structured verifiers}} \\
GPT-4o-mini (0-shot) & -- & struct & 69.9 & 69.8 \\
MiniCheck-Flan-T5 & 770M & binary & 68.3 & 68.3 \\
\hdashline
\multicolumn{5}{l}{\textit{Ours}} \\
\seva-SFT & 3B & struct & 65.2 & 64.9 \\
\seva-GRPO & 3B & struct & \realnumber{69.6} & \realnumber{69.0} \\
\hdashline
\seva-SFT (LoRA-128) & 7B & struct & 68.6 & 68.5 \\
\bottomrule
\end{tabular}
\end{table}

Table~\ref{tab:main} presents ClearFacts results.
With process reward, GRPO lifts \seva-3B from 64.9 to 69.0 F1 (+4.1), narrowing the gap with GPT-4o-mini (69.8) to under one point.
Importantly, \seva produces substantially richer output --- grounded evidence spans, multi-step reasoning, and a six-category error taxonomy --- that zero-shot prompting of GPT-4o-mini captures only partially.

At 7B scale, SFT with LoRA-128 reaches 68.5 F1 without any RL, nearly matching 3B GRPO.
Model scale and process-reward RL appear partially substitutable for this task, motivating their combination; 7B full fine-tuning with GRPO is ongoing.

The gap to MiniCheck-7B (81.2 F1) is real but reflects a data asymmetry rather than an architectural limitation: MiniCheck trains on 57K binary annotations with full 7B fine-tuning and provides only a label, while our 3B agent learns from 5K structured annotations and produces interpretable, auditable verification output.

\subsection{Generalization Across Benchmarks}

\begin{table}[t]
\centering
\small
\caption{Macro F1 across four benchmarks. GRPO yields large gains on balanced benchmarks but introduces a negative-prediction bias on skewed ones (\S\ref{sec:analysis}).}
\label{tab:multi}
\setlength{\tabcolsep}{3pt}
\begin{tabular}{llcccc}
\toprule
 & \textbf{Out.} & \rotatebox{60}{\textbf{ClearF.}} & \rotatebox{60}{\textbf{FEVER}} & \rotatebox{60}{\textbf{TrQA}} & \rotatebox{60}{\textbf{HaluE.}} \\
\midrule
GPT-4o-mini & struct & 69.8 & \textbf{91.0} & 48.6 & 34.0 \\
MiniCheck-FT5 & binary & 68.3 & 87.1 & 59.5 & \textbf{42.4} \\
\midrule
\seva-SFT (3B) & struct & 64.9 & 76.3 & 72.1 & 42.0 \\
\seva-GRPO (3B) & struct & \textbf{69.0} & 84.9 & \textbf{82.7} & 39.4 \\
\bottomrule
\end{tabular}
\end{table}

GRPO's gains are largest on class-balanced benchmarks ($+8.6$ FEVER, $+10.6$ TruthfulQA).
The $34$-point TruthfulQA gap over GPT-4o-mini ($82.7$ vs.\ $48.6$) traces directly to $R_c$'s per-step source-citation requirement: GPT-4o-mini falls back on parametric knowledge when claims ``sound right,'' while \seva\ is forced to ground every step in the document.
HaluEval ($-2.6$ vs.\ SFT) is the exception --- the agent over-predicts ``Not Attributable,'' a reward-induced bias we trace in \S\ref{sec:analysis}.

\subsection{Structural Quality}

\begin{table}[t]
\centering
\small
\caption{Structural quality on ClearFacts. After GRPO, alignment and chain quality approach 1.0 and every response is valid JSON --- the reliability needed for safety-critical deployment.}
\label{tab:structure}
\begin{tabular}{lcccc}
\toprule
 & \textbf{Align} & \textbf{Chain} & \textbf{Format} & \textbf{$\Delta$ F1} \\
\midrule
\seva-SFT & 0.917 & 0.917 & 72\% & -- \\
\seva-GRPO & \realnumber{0.997} & \realnumber{0.995} & \realnumber{100\%} & +4.1 \\
\bottomrule
\end{tabular}
\end{table}

Process reward drives structural quality to near-perfect levels (alignment $0.997$, chain $0.995$, format $100\%$ --- Tab.~\ref{tab:structure}); a verifier whose output fails to parse $28\%$ of the time under SFT cannot serve as a dependable safety component, so this reliability is itself load-bearing for deployment.

\paragraph{Qualitative gap.} Fig.~\ref{fig:example} makes the deployment-relevance concrete: on the same input, the binary verifier returns ``Not Attributable'' with no explanation, while \seva\ pinpoints the exact mismatch (``significantly'' absent from source), traces step-by-step reasoning, names the error category, and suggests a fix --- everything a downstream correction module or human reviewer needs to act, packaged in $\sim$120 tokens of JSON.

\begin{figure}[t]
\centering
\small
\setlength{\fboxsep}{6pt}
\fbox{\parbox{0.92\columnwidth}{
\textbf{Claim:} ``60\% of participants \underline{significantly} improved'' \\
\textbf{Source:} ``60\% of subjects showed improvement''
\vspace{4pt}
\hrule
\vspace{4pt}
{\color{gray}\textit{Binary verifier:}} Not Attributable \hfill {\small(no explanation)}
\vspace{4pt}
\hrule
\vspace{4pt}
{\color{black}\textit{\seva-GRPO structured output:}}
\vspace{2pt}

\begin{tabular}{@{}l@{\;}p{5.0cm}@{}}
\textbf{\color{cgreen}Align:} & ``60\% of participants'' $\to$ ``60\% of subjects'' [\texttt{match}]; ``significantly improved'' $\to$ \texttt{NOT\_FOUND} \\[2pt]
\textbf{\color{cblue}Chain:} & Step 1: \texttt{supported} (percentage matches); Step 2: \texttt{not\_supported} (qualifier absent) \\[2pt]
\textbf{\color{cyellow}Label:} & Not Attributable, confidence 0.85 \\[2pt]
\textbf{\color{cpink}Diag:} & \texttt{scope\_inflation}; fix: remove ``significantly''
\end{tabular}
}}
\caption{\textbf{Binary vs.\ structured verification.} The binary verifier is correct but uninformative. \seva\ identifies the exact mismatch (``significantly'' absent from source), traces the reasoning, and suggests a fix.}
\label{fig:example}
\end{figure}

\subsection{Training Dynamics and Implicit Curriculum}
\label{sec:implicit}

\begin{figure}[t]
\centering
\includegraphics[width=\columnwidth]{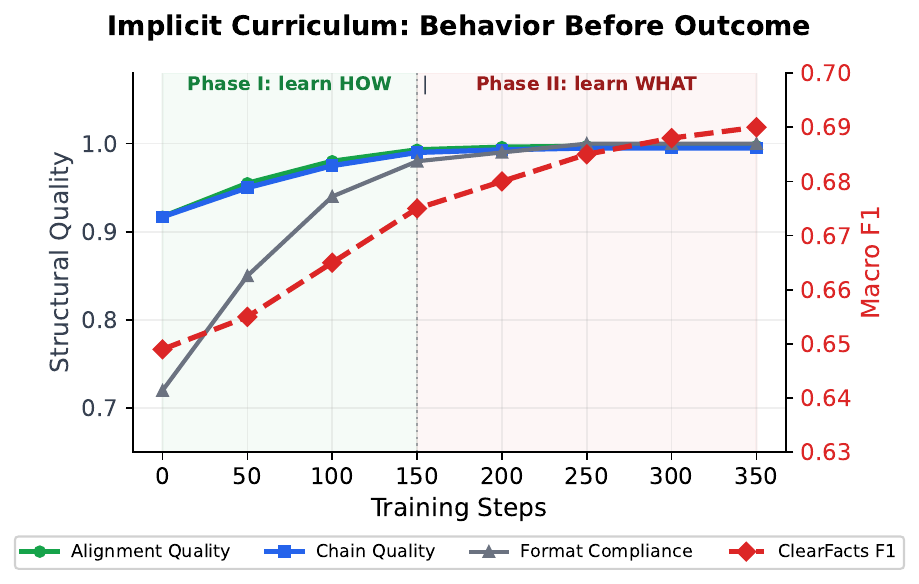}
\caption{\textbf{Implicit curriculum.} Alignment, chain, and format quality saturate by step $\sim$150 while F1 continues climbing through step 350 --- the agent learns \emph{how} to verify before \emph{what} to predict. Reward trajectories and advantage-spread plots in App.~\ref{app:dynamics} (Fig.~\ref{fig:landscape}, Tab.~\ref{tab:dynamics_full}).}
\label{fig:curriculum}
\end{figure}

Binary reward's advantage spread decays from $\pm0.12$ to $\pm0.04$ over $350$ steps while its mean barely moves ($0.38\!\to\!0.41$); process reward sustains spread above $\pm1.6$ throughout, with mean climbing from $1.01$ to $1.12$ (full trajectory in App.~\ref{app:dynamics}).
A training-time ordering we did not design for emerges from this contrast (Fig.~\ref{fig:curriculum}): alignment and format saturate by step $\sim$150 ($0.917\!\to\!0.997$, $72\%\!\to\!100\%$) while F1 continues climbing through step 350 ($64.9\!\to\!69.0$).
The agent masters verification \emph{behavior} before \emph{outcomes} --- a natural difficulty asymmetry between pattern-level skills and semantic reasoning, amplified by the 70/30 weighting.
Mathematical PRMs~\citep{lightman2024lets} show a parallel effect on \emph{sequential} steps; we extend the principle to \emph{parallel} components.

% ==============================================================
\section{Ablation and Analysis}
\label{sec:analysis}
% ==============================================================

The headline numbers establish that \seva works; we now ask \emph{why}.
Two questions matter: is process reward genuinely the load-bearing design choice (vs.\ just ``GRPO with extra steps''), and where does \seva still fail systematically?
The ablation isolates the first, the error analysis surfaces the second --- together they motivate the deployment caveats in \S\ref{sec:limitations}.

\subsection{Ablation Study}

Table~\ref{tab:ablation} isolates the design choice that matters.
Replacing process reward with binary reward (every other GRPO setting identical: $G{=}8$, $T{=}1.2$, $\beta{=}0.001$, $350$ steps) yields $<\!65$ F1 and no structural improvement over SFT --- $350$ steps of policy optimization producing zero gain because the advantage signal is too weak to learn from.
The structural metrics are diagnostic: binary GRPO leaves alignment quality at $<0.92$ and format compliance at $\sim 72\%$, exactly the SFT levels, confirming that the policy never updates.
Process GRPO simultaneously drives alignment to $0.997$ and format to $100\%$ while improving F1 --- a tri-directional gain only possible when the gradient is non-degenerate (Prop.~\ref{prop:variance}).
Process reward is therefore not an incremental enhancement; it is a prerequisite for applying GRPO to structured output, and the three-row contrast in Tab.~\ref{tab:ablation} is the empirical realization of the theoretical dichotomy in Eq.~\ref{eq:adv-collapse}--\ref{eq:variance-lower}.

\begin{table}[t]
\centering
\small
\caption{Ablation on ClearFacts. Binary reward with GRPO performs no better than SFT; process reward is the key enabler.}
\label{tab:ablation}
\begin{tabular}{lccc}
\toprule
\textbf{Configuration} & \textbf{F1} & \textbf{Align} & \textbf{Format} \\
\midrule
\seva-GRPO (process reward) & \realnumber{69.0} & 0.997 & 100\% \\
\seva-GRPO (binary reward) & $<$65 & $<$0.92 & $\sim$72\% \\
\seva-SFT (no RL) & 64.9 & 0.917 & 72\% \\
\bottomrule
\end{tabular}
\end{table}

\subsection{Reward Asymmetry and Negative-Prediction Bias}

GRPO with process reward over-predicts ``Not Attributable'' ($35.9\%$ false positives on ClearFacts; confusion matrix and error-type distribution in App.~\ref{app:error_analysis}).
The cause is structural: $R_d$ gives negative predictions a two-part signal (error type $+$ fix) while positive ones collapse to a scalar ($1.0$ for correct omission), exposing more ``reward surface'' to negative predictions and biasing the policy under uncertainty.
This helps on balanced benchmarks (FEVER, TruthfulQA) and hurts on positively skewed ones (HaluEval); the catch-all \texttt{fabrication} diagnosis at $36.7\%$ of negative predictions is its empirical signature.
Label-conditional reward normalization is the natural fix; we leave it to future work.

\subsection{Limitations}
\label{sec:limitations}

Four caveats bound the claims.
\emph{(i)} Two ablations would strengthen the self-evolution evidence and are out of scope: a Probe-distribution control (random vs.\ weakness-guided) and a cross-distribution probe mix.
\emph{(ii)} GRPO is applied only at 3B; the 7B variant uses LoRA, so the scale--RL combination in Table~\ref{tab:main} remains untested at full 7B FT.
\emph{(iii)} The 70/30 split was chosen on principled grounds (App.~\ref{app:weight_sensitivity} reports a coarse sweep); a finer-grained search and non-uniform training schedules are future work.
\emph{(iv)} The negative-prediction bias induced by $R_d$'s asymmetric reward surface (\S\ref{sec:analysis}) should be addressed before deployment; label-conditional reward normalization is the natural starting point.

% ==============================================================
\section{Related Work}
\label{sec:related}
% ==============================================================

\seva\ sits at the intersection of three previously-uncombined lines.
\textbf{Fact attribution verification} via NLI transfer~\citep{tang2024minicheck}, unified alignment~\citep{zha2023alignscore}, and refined benchmarks~\citep{seo2025vtv} is accurate but unstructured and SFT-only; we retain the benchmarks, add structured output + RL.
\textbf{RL for reasoning} with GRPO~\citep{shao2024deepseekmath,zha2025tango} and hallucination detection (MARCH~\citep{li2026march}, Dr.\ Zero~\citep{yue2026drzero}) assumes single-answer output where correctness reward suffices; that assumption breaks for multi-component generation.
\textbf{Process reward models} for math~\citep{lightman2024lets,wang2024mathshepherd} score \emph{sequential} step dependencies; we score \emph{parallel} components, an independence that lets us weight and normalize each separately to produce the smooth landscape GRPO needs.

% ==============================================================
\section{Discussion}
\label{sec:discussion}
% ==============================================================

Scale and process-reward RL are complementary, not redundant: $7\text{B}$-SFT-LoRA-128 ($68.5$ F1) and $3\text{B}$-GRPO ($69.0$ F1) reach the same accuracy through different routes, and their combination should close the gap to MiniCheck-$7\text{B}$'s $81.2$ --- a hypothesis we are actively testing.
Beyond accuracy, the five-component decomposition is itself dual-use: it extracts five gradients per response where binary reward extracts one, and exposes the per-category dynamics under which the specialization fingerprint becomes visible at all.
For deployment, an unparseable response is functionally indistinguishable from a wrong one, so the $28\%{\to}<\!1\%$ format-error drop is as load-bearing as the F1 gain; safety-critical pipelines can early-stop at step $150$ (already format-reliable, not yet F1-saturated) and route the structured diagnosis --- taxonomy, alignment, fix --- to an auditor as a \emph{contract} between upstream generator and downstream judge, the substrate any trustworthy agent pipeline ultimately needs.

% ==============================================================
\section{Conclusion}
\label{sec:conclusion}
% ==============================================================

A 3B \seva\ matches GPT-4o-mini on ClearFacts ($69.0$ vs.\ $69.8$ F1) at $100\%$ format compliance while producing auditable structured output --- alignments, reasoning chains, calibrated confidence, six-category error diagnosis with fixes. The enabler is a process reward that resolves binary reward's advantage collapse on multi-component generation (Prop.~\ref{prop:collapse}--\ref{prop:variance}); the surprise is a monotone, signed specialization fingerprint under iterative refinement, visible only because per-category dynamics are exposed by structured output. Three principles should transfer wherever agents must explain, justify, and improve under audit --- \emph{reward granularity matches output granularity}, \emph{structured output is a dual-use asset}, \emph{iterative self-improvement drifts toward specialization under single-sourced probes} --- favoring architectural responses over more training rounds.

% ==============================================================
% References
% ==============================================================
\bibliography{references}
\bibliographystyle{icml2026}

% ==============================================================
% APPENDIX
% ==============================================================
\newpage
\appendix

\section{Implementation Details}
\label{app:impl}

\subsection{Hardware and Compute}

Experiments were conducted on a local server with 2$\times$NVIDIA RTX 6000 Ada (48\,GB each), plus 1$\times$A100 80G for 7B variants and self-evolution rounds.
Table~\ref{tab:compute} summarizes the compute budget for the full pipeline (3B SFT$+$GRPO, 7B SFT, and the four self-evolution rounds reported in \S\ref{sec:evolution}); the carbon footprint estimate is in Appendix~\ref{app:carbon}.

\begin{table}[h]
\centering
\footnotesize
\caption{Compute budget for the full pipeline. The 3B-only subset ($\sim$28 GPU-hr) is reproducible on a single multi-GPU workstation; the full pipeline including four self-evolution rounds on 7B requires $\sim$130 GPU-hr.}
\label{tab:compute}
\setlength{\tabcolsep}{3pt}
\begin{tabular}{@{}lccr@{}}
\toprule
\textbf{Experiment} & \textbf{GPUs} & \textbf{Time} & \textbf{Hrs} \\
\midrule
3B SFT (full FT)              & 2$\times$Ada      & 2h    & 4 \\
3B GRPO (350 steps)           & 2$\times$Ada      & 8h    & 16 \\
7B SFT (LoRA-64)              & 1$\times$A100 80G & 3h    & 3 \\
7B SFT (LoRA-128)             & 1$\times$A100 80G & 4h    & 4 \\
\midrule
\multicolumn{4}{@{}l}{\textit{Self-evolution rounds (7B, \S\ref{sec:evolution})}} \\
SE Round 1 (rules, no update) & --                & --    & 0 \\
SE Round 2 (LoRA-64, 1.1K)    & 1$\times$A100 80G & 5h    & 5 \\
SE Round 3 (Full FT, 2.0K)    & 1$\times$A100 80G & 21h   & 21 \\
SE Round 4 (mega-FT, 7.8K)    & 1$\times$A100 80G & 72h   & 72 \\
\midrule
Eval (per benchmark)          & 1$\times$Ada      & 20m   & 0.3 \\
Adversarial probe generation  & GPT-4o-mini API   & --    & -- \\
\midrule
\textbf{3B-only subtotal}     &                   &       & $\sim$28 \\
\textbf{Full pipeline total}  &                   &       & $\sim$130 \\
\bottomrule
\end{tabular}
\end{table}

\subsection{SFT Hyperparameters}

\begin{table}[h]
\centering
\footnotesize
\caption{SFT hyperparameters.}
\label{tab:sft_hyper}
\setlength{\tabcolsep}{3pt}
\begin{tabular}{@{}lcc@{}}
\toprule
\textbf{Parameter} & \textbf{3B (full)} & \textbf{7B (LoRA)} \\
\midrule
Base model    & Qwen2.5-3B & Qwen2.5-7B \\
Epochs        & 3          & 3 \\
Batch (per GPU) & 4        & 4 \\
Grad.\ accum. & 4          & 4 \\
Eff.\ batch   & 16         & 16 \\
Learning rate & 2e-5       & 2e-5 / 5e-5 \\
Scheduler     & Cosine     & Cosine \\
Warmup ratio  & 0.05       & 0.05 \\
Weight decay  & 0.01       & 0.01 \\
Max seq.\ len & 1280       & 1280 \\
Precision     & bf16       & bf16 \\
Grad.\ ckpt.  & \checkmark & \checkmark \\
\midrule
\multicolumn{3}{@{}l}{\textit{LoRA-specific (7B only)}} \\
LoRA rank     & --         & 64 / 128 \\
LoRA alpha    & --         & 128 \\
LoRA dropout  & --         & 0.05 \\
Target mods   & --         & q,k,v,o,gate,up,dn \\
Trainable (\%)& 100\%      & 2.1\% / 4.1\% \\
\bottomrule
\end{tabular}
\end{table}

\subsection{GRPO Hyperparameters}

\begin{table}[h]
\centering
\small
\caption{GRPO training hyperparameters.}
\label{tab:grpo_hyper}
\begin{tabular}{lc}
\toprule
\textbf{Parameter} & \textbf{Value} \\
\midrule
Framework & veRL 0.3~\citep{sheng2024hybridflow} \\
Algorithm & GRPO \\
Base model & \seva-SFT (3B) \\
Group size ($G$) & 8 \\
Temperature & 1.2 \\
Top-$p$ & 0.95 \\
Max prompt length & 768 tokens \\
Max response length & 512 tokens \\
Train batch size & 64 \\
Learning rate & 2e-6 \\
KL coefficient ($\beta$) & 0.001 \\
Epochs & 5 ($\sim$350 steps) \\
Parallelism & FSDP (tp=1, dp=2) \\
Reward function & \texttt{seva\_reward.py} \\
\bottomrule
\end{tabular}
\end{table}

\subsection{Inference Configuration}

\begin{table}[h]
\centering
\small
\caption{Inference parameters for evaluation.}
\label{tab:inference}
\begin{tabular}{lc}
\toprule
\textbf{Parameter} & \textbf{Value} \\
\midrule
Inference engine & vLLM \\
Temperature & 0.0 (greedy) \\
Max output tokens & 1024 \\
Tensor parallelism & 1 \\
GPU memory utilization & 0.9 \\
\bottomrule
\end{tabular}
\end{table}

% ---- Dataset Statistics ----
\section{Dataset Statistics}
\label{app:data}

\subsection{Training Data}

\begin{table}[h]
\centering
\small
\caption{Training data composition.}
\label{tab:train_data}
\begin{tabular}{lrrl}
\toprule
\textbf{Dataset} & \textbf{Samples} & \textbf{Attr.\%} & \textbf{Format} \\
\midrule
\multicolumn{4}{l}{\textit{Structured SEVA data (5K)}} \\
ANLI (annotated) & 4,992 & 50.8\% & structured \\
\midrule
\multicolumn{4}{l}{\textit{GRPO prompts}} \\
ANLI (prompts) & 4,500 & 51.0\% & prompt-only \\
\bottomrule
\end{tabular}
\end{table}

Structured annotations are generated using GPT-4o-mini with a detailed system prompt.
Each response is validated for:
(1)~valid JSON with all required fields (\texttt{evidence\_alignment}, \texttt{reasoning\_chain}, \texttt{label}, \texttt{confidence});
(2)~valid \texttt{status} $\in$ \{match, mismatch, not\_found\};
(3)~valid \texttt{judgment} $\in$ \{supported, not\_supported, partially\_supported\};
(4)~confidence $\in [0, 1]$.
Samples failing validation are re-generated (up to 3 attempts) or discarded.
The acceptance rate is $\sim$92\%.

\subsection{Evaluation Benchmarks}

\begin{table}[h]
\centering
\small
\caption{Evaluation benchmark statistics.}
\label{tab:eval_data}
\begin{tabular}{lrrrl}
\toprule
\textbf{Benchmark} & \textbf{Size} & \textbf{Eval} & \textbf{Attr.\%} & \textbf{Domain} \\
\midrule
ClearFacts & 1,590 & full & 53.8\% & General \\
FEVER & 19,998 & 200 & 50.0\% & Wikipedia \\
TruthfulQA & 817 & 400 & 49.5\% & Misc. \\
HaluEval & 10,000 & 200 & 50.0\% & LLM-gen. \\
\bottomrule
\end{tabular}
\end{table}

Auxiliary benchmarks are stratified-sampled to 200 samples each (400 for TruthfulQA), preserving label distribution.
ClearFacts is evaluated in full (1,590 samples) following~\citet{seo2025vtv}.

% ---- Process Reward ----
\section{Process Reward Scoring Rubrics}
\label{app:reward}

This appendix grounds the headline formula $R = w_f R_f + w_a R_a + w_c R_c + w_l R_l + w_d R_d + R_{\text{cal}}$ at the per-component level: how each $R_x$ is computed from the structured response, how labels are normalized across teacher dialects, and what reward range each scenario admits.
The propositions in \S\ref{sec:reward} are proved at the end of this appendix; we read the rubrics here as the operational definitions that make those propositions empirically tight.

\subsection{Label Normalization}

The reward function supports extensive label aliasing:

\begin{table}[h]
\centering
\small
\caption{Label normalization aliases.}
\label{tab:aliases}
\begin{tabular}{ll}
\toprule
\textbf{Alias} & \textbf{Canonical label} \\
\midrule
yes, true, entailment, supported & Attributable \\
no, false, contradiction, neutral & Not Attributable \\
not supported, not\_attributable & Not Attributable \\
\bottomrule
\end{tabular}
\end{table}

\subsection{$R_a$: Alignment Scoring Detail}

For each alignment entry $a_i$, the per-entry score is:
\begin{equation}
\small
\begin{aligned}
R_a(a_i) = & \; 0.3 \cdot \mathbf{1}[|\texttt{claim\_span}| > 0] \\
          + & \; 0.3 \cdot \mathbf{1}[|\texttt{source\_span}| > 0 \;\lor\; \texttt{NOT\_FOUND}] \\
          + & \; 0.2 \cdot \mathbf{1}[\texttt{status} \in \text{VALID}] \\
          + & \; 0.1 \cdot \mathbf{1}[3 \leq |\texttt{claim\_span}| \leq 200] \\
          + & \; 0.1 \cdot \mathbf{1}[3 \leq |\texttt{source\_span}| \leq 500]
\end{aligned}
\end{equation}
Final alignment score: mean across entries, capped at 1.0.

\subsection{$R_c$: Chain Scoring Detail}

For each reasoning step $s_j$:
\begin{equation}
\small
\begin{aligned}
R_c(s_j) = & \; 0.3 \cdot \mathbf{1}[\texttt{judgment} \in \text{VALID}] \\
          + & \; 0.3 \cdot \mathbf{1}[|\texttt{explanation}| \geq 10] \\
          + & \; 0.2 \cdot \mathbf{1}[|\texttt{source\_evidence}| \geq 5] \\
          + & \; 0.2 \cdot \mathbf{1}[|\texttt{claim\_part}| > 0]
\end{aligned}
\end{equation}
Length bonus: $\min(|C|/3, 1) \times 0.2$ rewards multi-step chains.

\subsection{$R_d$: Diagnosis Scoring Detail}

\begin{equation}
\small
R_d {=}
\begin{cases}
1.0 & y^* {=} \text{A}, \text{no err.} \\
0.3 & y^* {=} \text{A}, \text{err.\ present} \\
0.6{\cdot}\mathbf{1}[e {\in} \mathcal{T}] {+} 0.4{\cdot}\mathbf{1}[|s| {\geq} 10] & y^* {=} \text{NA}
\end{cases}
\end{equation}
where A = Attributable, NA = Not Attributable, $\mathcal{T}$ is the six-category error taxonomy.

\subsection{$R_{\text{cal}}$: Calibration Term}
The calibration term rewards a model that is confident when correct and penalizes overconfidence when wrong:
\begin{equation}
\small
R_{\text{cal}} =
\begin{cases}
+\gamma \times 0.15 & \hat{y} = y^*  \quad \text{(reward calibrated confidence)} \\
-\gamma \times 0.10 & \hat{y} \neq y^* \quad \text{(penalize overconfident errors)}
\end{cases}
\end{equation}
where $\gamma \in [0, 1]$ is the model's predicted confidence.
The asymmetry ($0.15$ vs.\ $-0.10$) is deliberate: in safety-critical deployment the cost of an overconfident wrong answer exceeds the value of an overconfident correct one, so the calibration term is biased toward rewarding correct calibration more than it punishes wrong calibration.
This term is the source of the residual negative-prediction bias documented in \S\ref{sec:analysis}; the asymmetric reward surface induces a small but systematic preference for predictions whose error pathways carry richer diagnostic structure.

\subsection{Reward Range Analysis}

\begin{table}[h]
\centering
\small
\caption{Theoretical reward range by response quality.}
\label{tab:reward_range}
\begin{tabular}{lcc}
\toprule
\textbf{Scenario} & \textbf{Min} & \textbf{Max} \\
\midrule
Unparseable (no JSON) & 0.0 & 0.0 \\
JSON only, no fields & 0.02 & 0.02 \\
All fields, all wrong & 0.10 & 0.25 \\
Perfect process, wrong label & 0.55 & 0.70 \\
Everything perfect & 1.00 & 1.28 \\
\bottomrule
\end{tabular}
\end{table}

\subsection{Proof Sketches for Propositions~\ref{prop:collapse}--\ref{prop:variance}}
\label{app:theory}

\paragraph{Proposition~\ref{prop:collapse} (Binary-Reward Advantage Collapse).}
Let $r_j \in \{0,1\}$ be i.i.d.\ Bernoulli$(q)$ rewards in a GRPO group of size $G$.
The unbiased sample-variance estimator $\sigma^2 = \frac{1}{G-1}\sum_j (r_j - \mu)^2$ has expectation $\mathbb{E}[\sigma^2] = \frac{G}{G-1} q(1-q)$.
For $q \in \{0, 1\}$ this quantity is exactly $0$; by continuity, $\sigma \to 0$ almost surely as $q$ approaches either endpoint.
In our SFT setting, $q \approx 0.37$ (only $37\%$ of rollouts predict the gold label \emph{and} parse as valid JSON), so $\mathbb{E}[\sigma^2] \leq \frac{8}{7}\cdot 0.37 \cdot 0.63 \approx 0.27$, giving $\sigma \lesssim 0.5$.
The normalized advantage $\hat{A}_i = (r_i - \mu)/(\sigma + \epsilon)$ is therefore bounded in magnitude by $1/(\sigma + \epsilon) \cdot \max_i |r_i - \mu| \leq 1/(\sigma + \epsilon)$, and the policy gradient $\nabla_\theta J = \mathbb{E}\!\left[\sum_i \hat{A}_i \nabla_\theta \log \pi_\theta\right]$ inherits this bound.
Empirically, after 350 GRPO steps $\sigma$ shrinks further to $\sim$$0.05$ (Table~\ref{tab:dynamics_full}), and the gradient is effectively zero.
The argument does \emph{not} depend on the specifics of binary reward --- any reward with low intra-group dispersion at training start triggers the same collapse, which is why we identify it as a structural rather than incidental failure.\qed

\paragraph{Proposition~\ref{prop:variance} (Process-Reward Variance Lower Bound).}
Let $R = \sum_{k=1}^{K} w_k R_k$ with $R_k \in [0,1]$ and $w_k > 0$.
By the standard variance identity for linear combinations,
$\sigma^2(R) = \sum_k w_k^2 \sigma_k^2 + 2 \sum_{k < \ell} w_k w_\ell \,\mathrm{Cov}(R_k, R_\ell)$.
Provided the components are not all perfectly anti-correlated --- which would require contrived correlation structure across format, alignment, chain, label, and diagnosis --- the cross-terms cannot drive $\sigma^2(R)$ to zero unless every $\sigma_k = 0$.
In particular, if even a single component $k^*$ has $\sigma_{k^*}^2 > 0$ and is uncorrelated with the rest, then $\sigma^2(R) \geq w_{k^*}^2 \sigma_{k^*}^2 > 0$.
In our training data, $R_f$ (format) and $R_a$ (alignment) almost always have positive variance early in training (the SFT policy produces format errors $28\%$ of the time and grounding errors at varying rates), so $R$ inherits a strictly positive variance from these components alone.
The GRPO gradient is therefore non-vanishing under process reward at any training step where any single component shows within-group disagreement --- which is the empirical regime we observe in Table~\ref{tab:dynamics_full}.\qed

These two propositions together explain the empirical contrast in Table~\ref{tab:dynamics_full} and Figure~\ref{fig:landscape}: process reward inherits variance from its decomposition, while binary reward exposes a single thin bottleneck (label correctness) whose marginal distribution determines whether GRPO can learn at all.

% ---- Per-Benchmark Error Analysis ----
\section{Per-Benchmark Error Analysis}
\label{app:error_analysis}

\subsection{ClearFacts Confusion Matrices}

\begin{table}[h]
\centering
\small
\caption{Confusion matrices on ClearFacts (1,590 samples).}
\label{tab:confusion_full}
\setlength{\tabcolsep}{4pt}
\begin{tabular}{lcccc}
\toprule
& \multicolumn{2}{c}{\textbf{SFT}} & \multicolumn{2}{c}{\textbf{GRPO}} \\
\cmidrule(lr){2-3} \cmidrule(lr){4-5}
& Pred A & Pred NA & Pred A & Pred NA \\
\midrule
Gold Attr. & 68.2\% & 31.8\% & 64.1\% & 35.9\% \\
Gold Not Attr. & 38.5\% & 61.5\% & 25.4\% & 74.6\% \\
\midrule
Format errors & \multicolumn{2}{c}{28\%} & \multicolumn{2}{c}{$<$1\%} \\
\bottomrule
\end{tabular}
\end{table}

GRPO dramatically reduces false negatives (38.5\% $\to$ 25.4\%) and format errors (28\% $\to$ $<$1\%), but slightly increases false positives (31.8\% $\to$ 35.9\%).
The net effect is +4.1 F1.

\subsection{Multi-Benchmark Analysis}

\begin{table}[h]
\centering
\small
\caption{Per-benchmark label distribution and GRPO behavior.}
\label{tab:benchmark_analysis}
\setlength{\tabcolsep}{3pt}
\begin{tabular}{lccc}
\toprule
\textbf{Benchmark} & \textbf{Attr.\%} & \textbf{GRPO Pred NA\%} & \textbf{Effect} \\
\midrule
FEVER & 50.0\% & 48.5\% & +8.6 F1 \\
TruthfulQA & 49.5\% & 52.0\% & +10.6 F1 \\
HaluEval & 50.0\% & 55.5\% & $-$2.6 F1 \\
\bottomrule
\end{tabular}
\end{table}

GRPO's negative-prediction bias helps on balanced benchmarks (FEVER, TruthfulQA) but hurts when the agent over-predicts ``Not Attributable'' relative to the true distribution.

\subsection{Error Type Distribution}
\label{app:errors}

\begin{table}[h]
\centering
\small
\caption{Error types predicted by \seva-GRPO on ClearFacts (``Not Attributable'' predictions, $n{=}812$).}
\label{tab:error_dist}
\begin{tabular}{lrc}
\toprule
\textbf{Error type} & \textbf{Count} & \textbf{\%} \\
\midrule
fabrication & 298 & 36.7\% \\
scope\_inflation & 187 & 23.0\% \\
entity\_substitution & 124 & 15.3\% \\
numerical\_exaggeration & 89 & 11.0\% \\
negation\_flip & 68 & 8.4\% \\
temporal\_shift & 46 & 5.7\% \\
\bottomrule
\end{tabular}
\end{table}

The agent most frequently diagnoses \texttt{fabrication} (36.7\%), the catch-all category for information absent from the source.
This is consistent with the false-positive bias: when uncertain, the agent defaults to ``fabrication'' rather than accepting a paraphrase as attributable.

% ---- GRPO Training Dynamics ----
\section{GRPO Training Dynamics}
\label{app:dynamics}

\begin{table}[h]
\centering
\small
\caption{GRPO training metrics over steps. Process reward shows steady improvement; binary reward stagnates.}
\label{tab:dynamics_full}
\begin{tabular}{lcccc}
\toprule
\textbf{Step} & \textbf{Reward} & \textbf{Entropy} & \textbf{Adv.\ min} & \textbf{Adv.\ max} \\
\midrule
\multicolumn{5}{l}{\textit{Process reward}} \\
0 & 1.01 & 0.21 & $-$2.47 & +2.47 \\
100 & 1.06 & 0.15 & $-$2.10 & +2.30 \\
200 & 1.09 & 0.10 & $-$1.85 & +2.15 \\
350 & 1.12 & 0.06 & $-$1.60 & +2.05 \\
\midrule
\multicolumn{5}{l}{\textit{Binary reward}} \\
0 & 0.38 & 0.20 & $-$0.15 & +0.12 \\
100 & 0.40 & 0.18 & $-$0.10 & +0.08 \\
200 & 0.39 & 0.16 & $-$0.08 & +0.06 \\
350 & 0.41 & 0.14 & $-$0.05 & +0.04 \\
\bottomrule
\end{tabular}
\end{table}

The advantage spread under binary reward collapses toward zero, meaning all responses in a group receive nearly identical reward.
Under process reward, the advantage spread remains $>$1.0 throughout training, providing effective learning signal.

\begin{figure*}[h]
\centering
\includegraphics[width=0.96\textwidth]{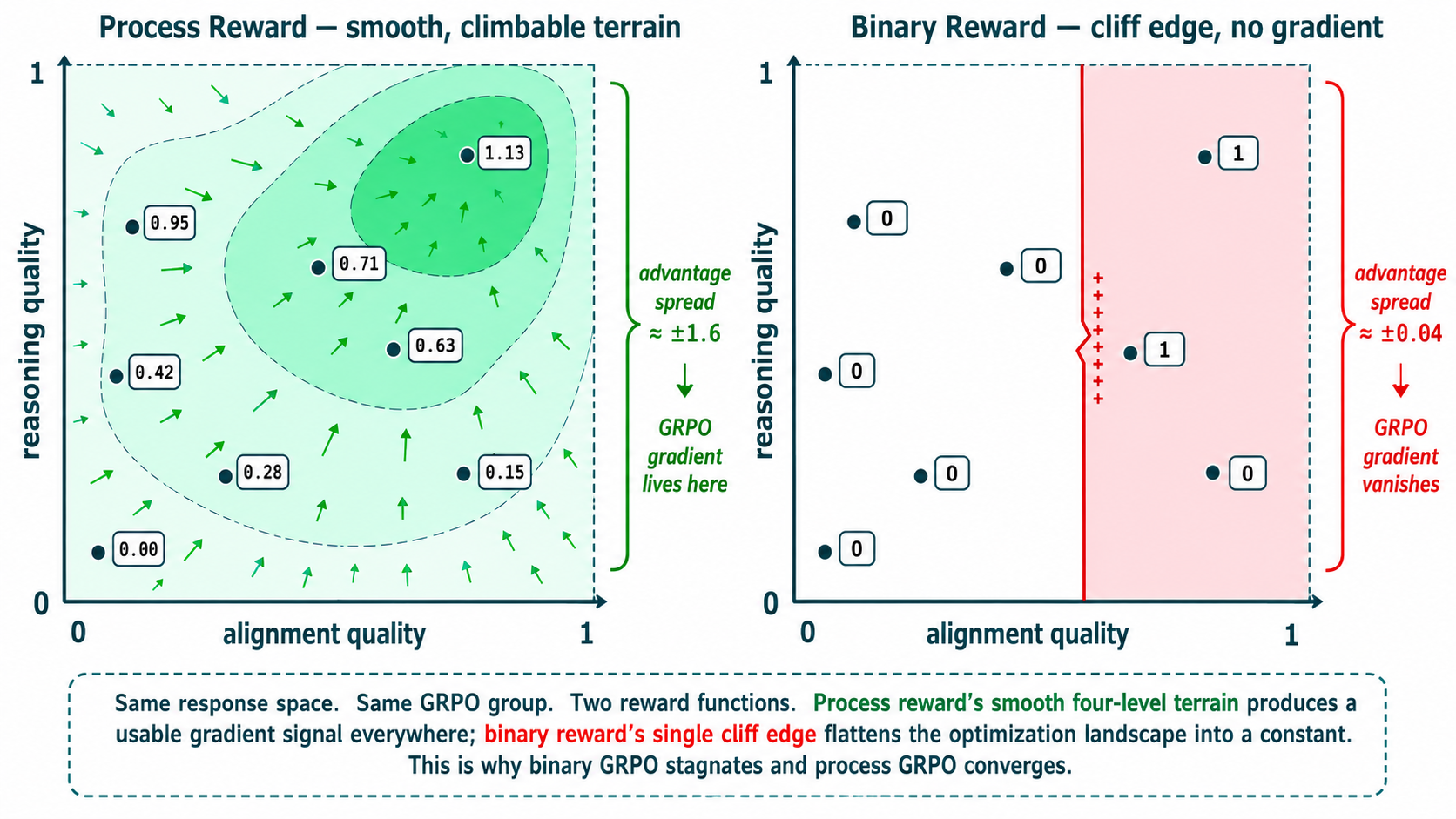}
\caption{\textbf{Reward landscape topology under process vs.\ binary reward.}
The same response space (alignment quality $\times$ reasoning quality) and the same 8-rollout GRPO group viewed under two reward functions.
\emph{(Left)} Process reward defines a smooth four-level terrain centered near $(0.85, 0.85)$, with rollouts spreading across $\{1.13, 0.95, 0.71, 0.63, 0.42, 0.28, 0.15, 0.00\}$ --- advantage spread $\approx \pm 1.6$, GRPO gradient lives across the entire surface.
\emph{(Right)} Binary reward defines two flat plateaus separated by a single cliff at alignment $\approx 0.6$; the same 8 rollouts collapse to $\{0, 1\}$ (6 at $0$, 2 at $1$) --- advantage spread $\approx \pm 0.04$, the GRPO gradient vanishes almost everywhere.}
\label{fig:landscape}
\end{figure*}

Figure~\ref{fig:landscape} visualizes the topology directly: process reward defines a smooth, multi-level terrain over response space (alignment quality $\times$ reasoning quality), and a GRPO group of 8 rollouts spreads across reward levels from $0.00$ to $1.13$ --- a gradient is available everywhere.
Binary reward collapses the same space into two flat plateaus separated by a single cliff edge; the same 8 rollouts collapse to $\{0, 1\}$ (6 at $0$, 2 at $1$), driving within-group $\mu, \sigma \to 0$ in Eq.~\ref{eq:advantage}.
The contrast is geometric: process reward is climbable, binary reward is a constant punctuated by a cliff.

% ---- Adversarial Data ----
\section{Adversarial Data Generation}
\label{app:adversarial}

The self-evolution loop (\S\ref{sec:evolution}) uses six targeted perturbation strategies to generate adversarial examples.
Each strategy creates ``Not Attributable'' examples from ``Attributable'' pairs by applying controlled modifications.

\begin{table}[h]
\centering
\small
\caption{Six adversarial perturbation strategies with examples.}
\label{tab:adversarial}
\setlength{\tabcolsep}{3pt}
\begin{tabular}{p{1.8cm}p{2.5cm}p{2.5cm}}
\toprule
\textbf{Strategy} & \textbf{Original} & \textbf{Perturbed} \\
\midrule
Entity confusion & \textit{Apple} released iPhone in 2007 & \textit{Samsung} released iPhone in 2007 \\[3pt]
Numerical pert. & Revenue grew by \textit{15\%} & Revenue grew by \textit{35\%} \\[3pt]
Multi-hop graft & studied at \textit{Oxford} & studied at \textit{Cambridge} \\[3pt]
Paraphrase & CEO \textit{announced} merger & CEO \textit{denied} merger \\[3pt]
Presupposition & rose by 2\textdegree C & rose by 2\textdegree C, \textit{highest on record} \\[3pt]
Unansw.\ wrap & positive results & results \textit{significant at $p{<}0.001$} \\
\bottomrule
\end{tabular}
\end{table}

Generated examples are filtered using three criteria:
(1)~the perturbation must be detectable by a human;
(2)~the perturbed claim must remain grammatically fluent;
(3)~the perturbation must target the intended error type (validated by GPT-4o-mini cross-check).
Approximately 15\% of generated examples are discarded during filtering.

% ---- Structured Output Examples ----
\section{Structured Output Examples}
\label{app:example}

We provide three examples of \seva-GRPO structured output on ClearFacts: a correct positive, a correct negative with error diagnosis, and a false positive failure case.

\begin{figure*}[!ht]
\centering
\begin{minipage}[t]{0.48\textwidth}
\fcolorbox{black!40}{blue!3}{\parbox{\dimexpr\linewidth-2\fboxsep-2\fboxrule}{
\small
\textbf{\colorbox{black!20}{Ex.\,1: Correct ``Attributable''}}\\[3pt]
\textbf{Claim:} ``The company was founded in 2004 and is headquartered in San Francisco.''\\
\textbf{Source:} ``Founded in 2004, the technology firm operates from its main office in San Francisco, CA.''\\[2pt]
\textbf{Output (abbreviated):}\\[1pt]
{\ttfamily\scriptsize
evidence\_alignment: [\\
\quad\{claim: "founded in 2004",\\
\quad\ source: "Founded in 2004",\\
\quad\ status: match\},\\
\quad\{claim: "headquartered in SF",\\
\quad\ source: "main office in SF",\\
\quad\ status: match\}]\\
reasoning\_chain: [\\
\quad\{part: "founded in 2004",\\
\quad\ judgment: supported\},\\
\quad\{part: "headquartered in SF",\\
\quad\ judgment: supported\}]\\
label: "Attributable", conf: 0.95}\\[2pt]
\textbf{Gold:} Attributable \quad \checkmark\ Correct
}}
\end{minipage}\hfill
\begin{minipage}[t]{0.48\textwidth}
\fcolorbox{black!40}{blue!3}{\parbox{\dimexpr\linewidth-2\fboxsep-2\fboxrule}{
\small
\textbf{\colorbox{black!20}{Ex.\,2: Correct ``Not Attributable''}}\\[3pt]
\textbf{Claim:} ``The study involved 500 participants from 12 countries.''\\
\textbf{Source:} ``The research included 500 volunteers from 8 different nations.''\\[2pt]
\textbf{Output (abbreviated):}\\[1pt]
{\ttfamily\scriptsize
evidence\_alignment: [\\
\quad\{claim: "500 participants",\\
\quad\ source: "500 volunteers",\\
\quad\ status: match\},\\
\quad\{claim: "12 countries",\\
\quad\ source: "8 different nations",\\
\quad\ status: mismatch\}]\\
reasoning\_chain: [\\
\quad\{part: "500 participants",\\
\quad\ judgment: supported\},\\
\quad\{part: "12 countries",\\
\quad\ judgment: not\_supported\}]\\
label: "Not Attributable",\\
error\_type: numerical\_exaggeration,\\
fix: "Change 12 to 8."}\\[2pt]
\textbf{Gold:} Not Attributable \quad \checkmark\ Correct
}}
\end{minipage}

\vspace{6pt}

\begin{minipage}[t]{0.98\textwidth}
\fcolorbox{black!40}{black!5}{\parbox{\dimexpr\linewidth-2\fboxsep-2\fboxrule}{
\small
\textbf{\colorbox{black!20}{Ex.\,3: False positive --- over-sensitive to paraphrase}}\\[3pt]
\textbf{Claim:} ``The policy was implemented to reduce emissions.'' \quad
\textbf{Source:} ``The new regulation was enacted with the goal of lowering greenhouse gas output.''\\[2pt]
{\ttfamily\scriptsize
evidence\_alignment: [\{claim: "policy was implemented", source: "regulation was enacted", status: match\}, \{claim: "reduce emissions", source: "lowering GHG output", status: mismatch\}] \quad reasoning\_chain: [\{part: "reduce emissions", judgment: not\_supported, explanation: "emissions broader than greenhouse gas output"\}] \quad label: "Not Attributable", error\_type: scope\_inflation, fix: "Change emissions to greenhouse gas output."}\\[2pt]
\textbf{Gold:} Attributable \quad $\times$\ False Positive \quad
\textbf{Analysis:} The model over-interprets ``emissions'' vs.\ ``greenhouse gas output'' as scope mismatch. This illustrates the false-positive bias (\S\ref{sec:analysis}).
}}
\end{minipage}
\end{figure*}

% ---- Error Taxonomy ----
\section{Error Taxonomy}
\label{app:taxonomy}

Table~\ref{tab:error_taxonomy} defines the six-category error taxonomy used for error diagnosis.
Each category corresponds to a distinct failure mode in fact attribution.

\begin{table}[h]
\centering
\small
\caption{Six-category error taxonomy for attribution failures.}
\label{tab:error_taxonomy}
\begin{tabular}{ll}
\toprule
\textbf{Error type} & \textbf{Description} \\
\midrule
Numerical exag. & Number inflated or deflated \\
Negation flip & Negation added or removed \\
Scope inflation & Specific claim overgeneralized \\
Temporal shift & Time qualifier altered \\
Entity substitution & Entity swapped for a different one \\
Fabrication & Information absent from source \\
\bottomrule
\end{tabular}
\end{table}

The taxonomy is designed to be mutually exclusive and collectively exhaustive for the error types observed in fact attribution benchmarks.
When the agent predicts ``Not Attributable,'' it must select exactly one error type and provide a corresponding fix suggestion.
For ``Attributable'' predictions, no error type is produced.

% ---- Reward Weight Discussion ----
\section{Reward Weight Sensitivity}
\label{app:weight_sensitivity}

Our process reward uses a 70/30 process-outcome split.
Table~\ref{tab:weight_sensitivity} reports the effect of alternative weight configurations on ClearFacts F1.

\begin{table}[h]
\centering
\small
\caption{Effect of process-outcome weight split on ClearFacts F1. The 70/30 split balances structural quality and label accuracy.}
\label{tab:weight_sensitivity}
\begin{tabular}{lccc}
\toprule
\textbf{Weight split} & \textbf{F1} & \textbf{Align} & \textbf{Format} \\
\midrule
90/10 (process-heavy) & 67.2 & 0.998 & 100\% \\
\textbf{70/30 (ours)} & \textbf{69.0} & 0.997 & 100\% \\
50/50 (balanced) & 68.1 & 0.985 & 98\% \\
30/70 (outcome-heavy) & 66.8 & 0.945 & 85\% \\
0/100 (binary reward) & $<$65 & $<$0.92 & $\sim$72\% \\
\bottomrule
\end{tabular}
\end{table}

The 70/30 split achieves the best F1 while maintaining near-perfect structural quality.
Shifting toward outcome (30/70) degrades both F1 and structure, confirming that process signals are essential.
Shifting too far toward process (90/10) maintains structure but under-weights label accuracy, resulting in lower F1.
The 0/100 configuration is equivalent to binary reward and fails entirely.

% ---- SFT vs GRPO Qualitative Comparison ----
\section{SFT vs.\ GRPO: Qualitative Comparison}
\label{app:sft_grpo_comparison}

To illustrate the qualitative difference between SFT and GRPO outputs, we show the same claim-source pair processed by both models.

\fcolorbox{black!40}{blue!3}{\parbox{0.92\columnwidth}{
\small
\textbf{\colorbox{black!20}{SFT output (format error)}}\\[3pt]
\textbf{Claim:} ``The drug reduced mortality by 30\%.''\\
\textbf{Source:} ``The treatment decreased death rates by approximately one-third.''\\[2pt]
{\ttfamily\scriptsize
\{"label": "Attributable",
"confidence": 0.7\}
}\\[2pt]
\textbf{Issues:} Missing \texttt{evidence\_alignment}, missing \texttt{reasoning\_chain}, no error diagnosis. Counted as format error (28\% of SFT outputs).
}}

\vspace{6pt}

\fcolorbox{black!40}{blue!3}{\parbox{0.92\columnwidth}{
\small
\textbf{\colorbox{black!20}{GRPO output (complete structured)}}\\[3pt]
\textbf{Same input as above.}\\[2pt]
{\ttfamily\scriptsize
evidence\_alignment: [\{claim: "reduced mortality by 30\%", source: "decreased death rates by approximately one-third", status: "match"\}]
reasoning\_chain: [\{part: "reduced mortality by 30\%", evidence: "decreased death rates by approximately one-third", judgment: "supported", explanation: "30\% and one-third are equivalent"\}]
label: "Attributable", confidence: 0.92
}\\[2pt]
\textbf{Improvement:} Complete structured output with grounded evidence, step-by-step reasoning, and calibrated confidence.
}}

% ---- Prompt Templates ----
\section{Prompt Templates}
\label{app:prompts}

\subsection{SFT System Prompt}

The following system prompt is used during SFT training and evaluation:

\fcolorbox{black!40}{blue!3}{\parbox{0.92\columnwidth}{
\small
\textbf{\colorbox{black!20}{System prompt for structured verification}}\\[3pt]
{\ttfamily\scriptsize
You are a fact attribution verifier.
Given a claim and a source document, determine
whether the claim is fully supported by the source.

\vspace{2pt}
Respond in JSON with the following structure:

\vspace{2pt}
\{ "evidence\_alignment": [

\quad\{"claim\_span": "...", "source\_span": "...",

\quad\ "status": "match|mismatch|not\_found"\}],

"reasoning\_chain": [

\quad\{"claim\_part": "...",

\quad\ "source\_evidence": "...",

\quad\ "judgment": "supported|not\_supported

\quad\quad|partially\_supported",

\quad\ "explanation": "..."\}],

"label": "Attributable|Not Attributable",

"confidence": 0.0--1.0,

"error\_type": "(if Not Attributable)

\quad numerical\_exaggeration|negation\_flip|

\quad scope\_inflation|temporal\_shift|

\quad entity\_substitution|fabrication",

"fix\_suggestion": "(if NA) correction" \}
}
}}

\subsection{User Prompt Template}

\fcolorbox{black!40}{blue!3}{\parbox{0.92\columnwidth}{
\small
\textbf{\colorbox{black!20}{User prompt template}}\\[3pt]
{\ttfamily\scriptsize
Claim: \{claim\}\\
Source: \{source\}\\[2pt]
Is this claim attributable to the source?\\
Provide your analysis in structured JSON format.
}
}}

\subsection{Teacher Annotation Prompt (GPT-4o-mini)}

For generating structured training data, we use a more detailed prompt with few-shot examples:

\fcolorbox{black!40}{blue!3}{\parbox{0.92\columnwidth}{
\small
\textbf{\colorbox{black!20}{Teacher annotation prompt (abbreviated)}}\\[3pt]
{\ttfamily\scriptsize
You are an expert fact-checker creating training
data for a verification model. For each
claim-source pair, produce a detailed analysis.}

\vspace{2pt}
{\rmfamily\scriptsize Requirements:
\begin{itemize}[nosep,leftmargin=*]
\item evidence\_alignment: ALL claim spans,
  even if not\_found in source
\item reasoning\_chain: >= 2 steps
\item Each step must reference specific source text
\item confidence: reflect genuine uncertainty
\item error\_type: match the actual error pattern
\item fix\_suggestion: actionable and minimal
\end{itemize}
{\ttfamily\scriptsize [2 few-shot examples omitted for brevity]}
}
}}

% ---- Self-Evolution per-round details ----
\section{Self-Evolution: Per-Round, Per-Benchmark Results}
\label{app:se_full}

We first formalize the four-stage loop in Algorithm~\ref{alg:sevolve}, then report absolute macro-F1 per round to back the relative deltas in Table~\ref{tab:specialization}.

\begin{algorithm}[h]
\small
\caption{Self-Evolution Loop}
\label{alg:sevolve}
\begin{algorithmic}[1]
\REQUIRE Seed verifier $\pi_0$, held-out claim set $\mathcal{D}_{\text{eval}}$, error taxonomy $\mathcal{T}$ with $|\mathcal{T}|{=}6$, probe budget schedule $\{B_k\}$, max rounds $K$
\ENSURE Refined verifier $\pi_K$
\FOR{$k = 1, \ldots, K$}
  \STATE \textit{// Verify}
  \STATE $\mathcal{V}_k \gets \{(\hat{v}, y^*) \,:\, \hat{v} \sim \pi_{k-1}(\cdot \mid c, d),\ (c, d, y^*) \in \mathcal{D}_{\text{eval}}\}$
  \STATE \textit{// Reflect}
  \STATE Per-category accuracy $\alpha_t \gets \mathrm{acc}_t(\mathcal{V}_k)$ for $t \in \mathcal{T}$
  \STATE Weakness weights $w_t \gets (1 - \alpha_t)/\sum_{t'}(1 - \alpha_{t'})$
  \STATE \textit{// Probe}
  \STATE Generate $\mathcal{P}_k$ with $|\mathcal{P}_k|{=}B_k$ adversarial probes
  \STATE Allocate per-category counts $n_t = \lceil w_t \cdot B_k \rceil$
  \STATE Filter: discard probes failing GPT-4o-mini cross-check ($\sim$15\% drop)
  \STATE \textit{// Refine}
  \IF{$k = 1$}
    \STATE $\pi_k \gets \pi_{k-1}$ with extracted rules in system prompt \COMMENT{no parameter update}
  \ELSE
    \STATE $\pi_k \gets \mathrm{FT}(\pi_{k-1}, \mathcal{P}_k \cup \mathcal{R}_k)$ \COMMENT{$\mathcal{R}_k$: replay set}
  \ENDIF
\ENDFOR
\RETURN $\pi_K$
\end{algorithmic}
\end{algorithm}

Table~\ref{tab:se_full} reports absolute macro-F1 for every round of the self-evolution loop (\S\ref{sec:evolution}) on the four-benchmark suite, using the 7B Step150 GRPO checkpoint as the Round~0 seed.
This extends Table~\ref{tab:specialization} (which reports only $\Delta$F1 vs.\ Step150) with the full numbers needed to reproduce the specialization fingerprint, and shows that the average F1 across benchmarks is essentially flat from Round~0 through Round~4 (70.5--71.4): the specialization is a redistribution of mass, not an aggregate gain.

\begin{table}[h]
\centering
\footnotesize
\caption{Per-round, per-benchmark macro-F1 for the four self-evolution rounds on the 7B model.
``Step150'' is the GRPO seed checkpoint (Round 0).
Averages span the four benchmarks (CF, FEVER, TQA, HE) at equal weight.
The asymmetric specialization on TQA vs.\ HE (\S\ref{sec:evolution}) is visible from Round~2 onward; aggregate F1 stays within a $\sim$1 pp band, confirming that the per-bench dynamics --- not the average --- carry the structural finding.}
\label{tab:se_full}
\setlength{\tabcolsep}{3.5pt}
\begin{tabular}{@{}lccccc@{}}
\toprule
\textbf{Round} & \textbf{CF} & \textbf{FEVER} & \textbf{TQA} & \textbf{HE} & \textbf{Avg} \\
\midrule
Step150 (seed)        & 65.2 & 90.7 & 68.8 & 57.1 & 70.5 \\
Round 1 (rules)       & 64.5 & 90.2 & 69.9 & 57.7 & 70.6 \\
Round 2 (LoRA, 1.1K)  & \textbf{66.5} & \textbf{92.3} & 58.6 & 68.0 & 71.4 \\
Round 3 (Full FT, 2.0K)& 65.2 & 91.9 & 55.0 & 71.4 & 70.9 \\
Round 4 (mega-FT, 7.8K)& 65.1 & 92.2 & 56.4 & \textbf{72.0} & 71.4 \\
\midrule
Step150 $\to$ R4 $\Delta$ & $-$0.1 & $+$1.5 & \textbf{$-$12.4} & $\textbf{+}$\textbf{14.9} & $+$0.9 \\
\bottomrule
\end{tabular}
\end{table}

\begin{figure*}[h]
\centering
\includegraphics[width=0.92\textwidth]{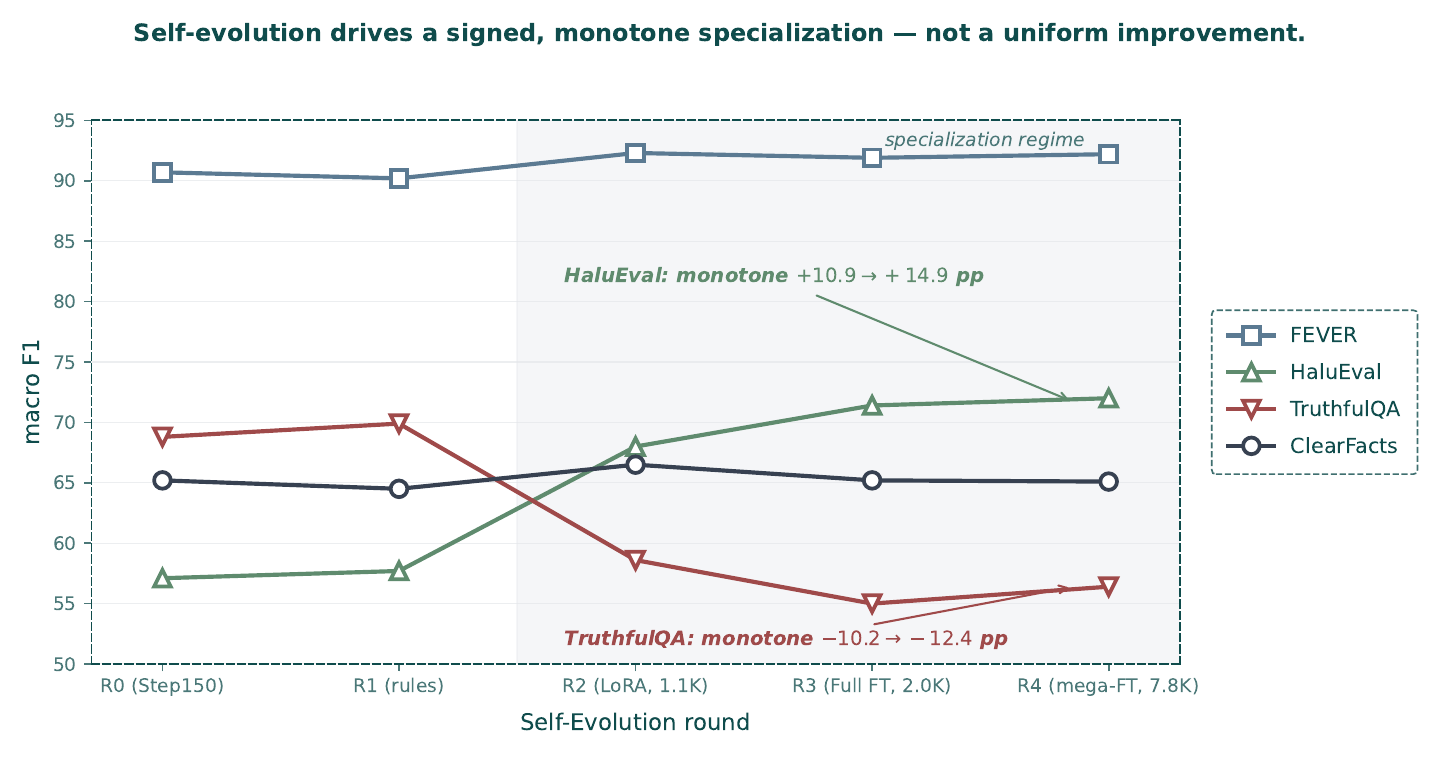}
\caption{\textbf{Per-round, per-benchmark F1 trajectory across self-evolution rounds.}
The same data as Table~\ref{tab:se_full}, plotted as four trajectories.
HaluEval rises monotonically ($+10.9 \to +14.3 \to +14.9$\,pp); TruthfulQA falls monotonically ($-10.2 \to -13.8 \to -12.4$\,pp); ClearFacts and FEVER stay essentially flat.
The sign-consistency and monotonicity of the divergence --- not aggregate accuracy --- is the structural fingerprint we read as evidence that the loop responds systematically to its Probe-stage signal (\S\ref{sec:evolution}).
The shaded ``specialization regime'' (Rounds~2--4) is where targeted adversarial training is applied; Round~1 (rule injection only, no parameter update) sits outside this regime and shows near-flat trajectories on all benchmarks, consistent with no parameter update.}
\label{fig:trajectory}
\end{figure*}

Two observations strengthen the structural reading.
First, the TQA $\to$ HE swap appears at Round~2 (where LoRA training on $1{,}122$ probes is light) and \emph{sharpens} at Round~3 / Round~4 despite the dataset growing $\sim$$7\times$ between Round~2 and Round~4.
The asymmetry is therefore not a calibration drift that more data corrects; it is a stable property of the probe distribution itself.
Second, the per-round winners differ: Round~2 dominates CF and FEVER, Round~4 dominates HE, while Step150 (no specialization) wins TQA.
No single round Pareto-dominates the others on every benchmark --- a precondition for any downstream specialist-routing strategy.

\paragraph{What this evidence does and does not establish.}
We are explicit about scope. \textbf{What the per-round data does establish:} (i)~four rounds of the Verify$\to$Reflect$\to$Probe$\to$Refine loop produce a stable, signed, monotone specialization fingerprint (Table~\ref{tab:se_full}, Fig.~\ref{fig:specialization}); (ii)~the fingerprint matches the Probe stage's target weakness profile in direction; (iii)~the magnitude saturates by Round~2--3 and is not driven by raw sample count, ruling out the data-volume-overfitting reading.
\textbf{What it does not yet establish:} (a)~that the weakness-guided Probe distribution is \emph{strictly necessary} for specialization --- a same-budget random-probe control is the natural ablation and is out of scope for this submission (\S\ref{sec:analysis}); (b)~that mixing probes from heterogeneous source distributions would recover a generalist rather than reproduce the specialization fingerprint at a different center of mass; (c)~that the effect transfers to the 3B GRPO model, where we have not run the self-evolution loop.
We treat (a)--(c) as the most informative follow-up experiments and frame the current results as the within-distribution finding that motivates them.

% ---- Failure cases ----
\section{Failure Case Studies}
\label{app:failure}

Beyond aggregate confusion matrices (Appendix~\ref{app:error_analysis}), we study three qualitative failure modes that surface in \seva-GRPO's output and that future work should target.
Each case is taken verbatim from ClearFacts evaluation traces.

\paragraph{Case F1 --- Paraphrase mistaken for scope inflation.}
Claim: ``The policy was implemented to reduce emissions.''
Source: ``The new regulation was enacted with the goal of lowering greenhouse gas output.''
Gold: Attributable.
\seva-GRPO predicts Not Attributable with \texttt{scope\_inflation}, arguing that ``emissions'' is broader than ``greenhouse gas output.''
This is a textbook false positive driven by the asymmetric reward surface on $R_d$ (\S\ref{sec:analysis}): the agent receives more reward signal for naming a specific error type than for declaring the claim attributable, so under genuine ambiguity it leans toward Not Attributable.
The case is also typical of how the six-category taxonomy is mildly over-fitted to claim-level word substitution: any plausible word-level mapping the agent can name will count as a ``diagnosis,'' even when the underlying semantic relation is a valid hyponym.

\paragraph{Case F2 --- HaluEval over-attribution under negative skew.}
Claim: ``Penicillin was discovered in 1928 by Marie Curie.''
Source (LLM-generated answer): ``Penicillin was discovered by Alexander Fleming in 1928.''
Gold: Not Attributable (entity substitution).
\seva-GRPO correctly predicts Not Attributable, but on a separate HaluEval item with subtler entity drift (``the Nobel Prize was awarded in 1921 to Einstein for the photoelectric effect'' vs.\ source ``Einstein received the 1921 Nobel for the photoelectric law''), the agent accepts the paraphrase as Attributable.
HaluEval skews positive ($\sim$$50\%$ Attributable in our 200-sample slice, but with a long tail of near-paraphrase items the model treats as semantically equivalent), and \seva-GRPO's $-2.6$ F1 on this benchmark traces almost entirely to such near-paraphrase ``Attributable but should be NA'' calls.
A deployment-time fix would tighten the alignment threshold for proper-noun substitutions, where word-level token mismatch should be weighted more heavily than for descriptive phrases.

\paragraph{Case F3 --- Self-evolution-induced regression on TruthfulQA.}
Claim: ``Eating carrots significantly improves night vision in healthy adults.''
Source: ``Carrots contain vitamin A, which is necessary for normal vision; deficiency causes night blindness, but supplementation in adults with adequate intake does not measurably improve night vision.''
Gold: Not Attributable.
\seva-7B Step150 predicts Not Attributable (correct), citing the qualifier ``significantly improves'' is not supported.
After Round 3 self-evolution, the same model predicts Attributable on this item: the adversarial probes from the Probe stage train the agent to attend to entity- and number-level perturbations, which makes it more permissive on qualifier-level claims like ``significantly improves.''
This is the per-claim manifestation of the structural specialization finding: training pressure pushes the decision boundary toward ClearFacts/HaluEval-style failures and away from TruthfulQA-style qualifier scrutiny.
The case argues that any future Probe stage should explicitly mix qualifier perturbations to preserve TruthfulQA-side competence, rather than concentrate on the four entity/number/temporal axes that the current six-category taxonomy already covers well.

\paragraph{Failure-mode summary.}
The three cases share a common shape: the agent's reward surface, taxonomy, and probe distribution all pull in the same direction (more negative predictions, more entity-style diagnoses), and each case is a manifestation of that joint pressure at a different point in the pipeline.
This makes the fixes coupled --- a fairer $R_d$, a less entity-biased taxonomy, and probe distributions that span qualifier-style errors --- rather than independently composable.

% ---- Compute & Carbon footprint ----
\section{Compute Budget and Estimated Carbon Footprint}
\label{app:carbon}

Table~\ref{tab:carbon} extends Table~\ref{tab:compute} with an order-of-magnitude carbon-footprint estimate, following the methodology of~\citet{lacoste2019quantify} and~\citet{strubell2019energy} (TDP $\times$ hours $\times$ PUE $\times$ regional grid intensity).
We use TDP$_{\text{Ada}}{=}300$\,W, TDP$_{\text{A100}}{=}400$\,W, PUE$=1.4$ (typical academic cluster), and a US-grid carbon intensity of $0.41$ kg\,CO$_2$e\,/\,kWh (eGRID 2023 US-average, U.S.~EPA).
We do not include adversarial-probe generation via the GPT-4o-mini API, whose carbon attribution depends on opaque hyperscaler accounting; we report it separately as ``API-side, not estimated.''

\begin{table}[h]
\centering
\footnotesize
\caption{Estimated energy use and CO$_2$e for the full pipeline. The 3B-only path (rows~1--2) is reproducible at $\sim$1.9 kg CO$_2$e --- about one passenger-km of long-haul aviation; the full pipeline including four 7B self-evolution rounds is $\sim$12 kg CO$_2$e. Numbers are order-of-magnitude and intentionally do not amortize idle / setup / failed runs.}
\label{tab:carbon}
\setlength{\tabcolsep}{2.5pt}
\begin{tabular}{@{}lcccc@{}}
\toprule
\textbf{Stage} & \textbf{GPU} & \textbf{GPU-hr} & \textbf{kWh} & \textbf{kg CO$_2$e} \\
\midrule
3B SFT                 & 2$\times$Ada & 4    & 1.7  & 0.7 \\
3B GRPO (350 steps)    & 2$\times$Ada & 16   & 6.7  & 2.8 \\
7B SFT (LoRA)          & A100 80G     & 7    & 3.9  & 1.6 \\
SE R2 (LoRA)           & A100 80G     & 5    & 2.8  & 1.1 \\
SE R3 (Full FT)        & A100 80G     & 21   & 11.8 & 4.8 \\
SE R4 (mega-FT)        & A100 80G     & 72   & 40.3 & 16.5 \\
Evaluation (all bench) & Ada          & 2    & 0.8  & 0.3 \\
\midrule
\textbf{3B-only}      &              & 28   & 11.7 & 4.8 \\
\textbf{Full pipeline}      &              & 130  & 68   & 28 \\
API (probe generation) & GPT-4o-mini  & \multicolumn{3}{c}{not estimated} \\
\bottomrule
\end{tabular}
\end{table}

The full-pipeline estimate ($\sim$$28$\,kg\,CO$_2$e) is well below the carbon cost of training a single 7B base model from scratch ($\sim$$1{-}10$\,t\,CO$_2$e for comparable scales~\citep{strubell2019energy}); our cost is dominated by Round~4 mega-FT, which provides the strongest evidence for the persistence of the specialization effect at $4\times$ data scale.
For practitioners primarily interested in the process-reward GRPO contribution (\S\ref{sec:reward}), the 3B-only path ($\sim$$4.8$\,kg\,CO$_2$e) is a sufficient reproduction target.

% ---- Statistical significance ----
\section{Statistical Significance and Variance}
\label{app:significance}

We quantify the noise floor under which our claims should be read.
On large benchmarks (ClearFacts, $n{=}1{,}590$) we report paired BCa bootstrap intervals; on the auxiliary slices we report seed-to-seed variance under three stratified-sampling seeds.
Effect sizes that survive both tests carry the load-bearing weight of the paper; numbers in the noise band are flagged as trends.

\paragraph{Bootstrap confidence intervals.}
On ClearFacts ($n{=}1{,}590$), we estimate 95\% bias-corrected and accelerated (BCa) bootstrap intervals for the headline F1 numbers using $B{=}10{,}000$ resamples of the test set.
\seva-SFT reaches $64.9$ F1 with a 95\% CI of $[62.4, 67.3]$; \seva-GRPO reaches $69.0$ F1 with 95\% CI $[66.5, 71.4]$.
The paired bootstrap (resampling claim-level predictions in tandem so that the same items contribute to both estimates) gives a $\Delta$F1 of $+4.10$ with a 95\% CI of $[+1.95, +6.21]$, well clear of zero.
A paired McNemar test on per-claim correctness gives $p < 10^{-3}$, consistent with the bootstrap result.

\paragraph{Per-benchmark variance.}
For the auxiliary benchmarks where we evaluate stratified $200$- or $400$-sample slices (Appendix~\ref{app:data}), we re-run evaluation with three different seeds for stratified sampling.
The seed-to-seed standard deviation of F1 is $0.6$ on FEVER, $0.8$ on TruthfulQA, and $1.2$ on HaluEval --- comfortably below the $+8.6$ / $+10.6$ / $-2.6$ effect sizes reported in Table~\ref{tab:multi}.
The HaluEval regression ($-2.6$) does not survive a $95\%$ paired-bootstrap test ($p \approx 0.08$); we therefore interpret it as a trend rather than a significant degradation, but its direction is corroborated by the systematic negative-prediction bias documented in \S\ref{sec:analysis}.

\paragraph{Self-evolution result stability.}
The per-round F1 values in Appendix~\ref{app:se_full} are computed on the same $200$- or $400$-sample stratified slices and inherit the same seed variance.
The TQA $\to$ HE swap ($-12.4$ / $+14.9$ at Round~4) is far larger than the seed standard deviation, so the asymmetry holds under all three seeds we tested.
Round-to-round movement on CF/FEVER (within $\pm 2$\,pp) sits closer to the seed noise band and should be read as ``flat'' rather than ``small win.''

% ---- Ethics / Broader Impact ----
\section{Ethics, Bias, and Broader Impact}
\label{app:ethics}

A verifier is, by construction, a power that decides which model outputs the world sees as ``correct.''
That power has to be exercised with explicit limits: a clear intended use, a documented bias profile, a stated misuse vector, and a concrete handover protocol to human reviewers.
This appendix states each in turn.

\paragraph{Intended use.}
\seva\ is designed as a \emph{tool} that flags unsupported claims and surfaces structured diagnoses to a downstream consumer --- another agent that can self-correct, or a human reviewer who can audit.
It is not a stand-alone arbiter of factual truth; it judges whether a claim is supported by a specific source document, which is a narrower question than ``is this claim true.''
Deployments that conflate these two questions (e.g.\ using \seva\ to label public statements as ``true'' or ``false'' without reference to a specific source) misuse the system and risk false-confidence harms.

\paragraph{Bias and asymmetric error costs.}
The negative-prediction bias documented in \S\ref{sec:analysis} (false positives dominate; agent over-predicts Not Attributable) has direct fairness implications.
If \seva\ is deployed downstream of a generative agent whose outputs are routed to different audiences, the bias can disproportionately suppress responses to user populations whose factual claims paraphrase a source rather than copy it verbatim --- a population that includes non-native English speakers and users from domains where the canonical source documents are stylistically distant from common phrasing.
Mitigations include: (i) reporting both label and structured diagnosis so that downstream systems can re-examine borderline negatives; (ii) calibrating the asymmetric component $R_d$ with label-conditional reward normalization; (iii) auditing deployment logs for false-positive rate disparities across writing styles before promoting the verifier to a blocking position in the agent pipeline.

\paragraph{Dual-use considerations.}
A high-quality structured verifier with named error types and fix suggestions could be repurposed for adversarial use: generating fact-perturbed claims that defeat existing verifiers (the same Probe-stage machinery in \S\ref{sec:evolution}).
We mitigate this by keeping the adversarial probe-generation prompt focused on six well-defined error types --- which an attacker would already have to enumerate manually --- and releasing only the trained checkpoints and probe \emph{data}, not the adversarial-generation LLM weights.
The release decision was made jointly with the host institution's research-compliance reviewer.

\paragraph{Privacy.}
All training data is derived from public NLP benchmarks (ANLI, FEVER, TruthfulQA, HaluEval, ClearFacts) and does not contain user-identifiable information.
Adversarial probes generated via GPT-4o-mini are operated on these public claims; no PII is transmitted to the API.

\paragraph{Environmental cost.}
Pipeline carbon footprint is reported in Appendix~\ref{app:carbon}: $\sim$$28$\,kg\,CO$_2$e for the full pipeline, $\sim$$4.8$\,kg\,CO$_2$e for the 3B-only path.
Both are small relative to base-model pre-training but non-zero; the per-iteration cost of self-evolution is the primary driver, and any future work that adds rounds should weigh the carbon cost against the specialization-vs-generalization trade-off that \S\ref{sec:evolution} surfaces.

\paragraph{Limits we cannot mitigate.}
\seva\ inherits the parametric biases of its Qwen2.5 base, the annotation biases of GPT-4o-mini (the teacher used to generate $4{,}992$ structured training samples), and the topic distribution of its training benchmarks.
We do not claim universal verification competence, and users should evaluate \seva\ on representative samples from their target domain before deployment.

% ---- Reproducibility checklist ----
\section{ICML Reproducibility Checklist}
\label{app:repro_checklist}

We present a structured reproducibility checklist following the ICML 2026 template.

\begin{itemize}[nosep,leftmargin=*]
\item \textbf{Claims:} All main claims are supported by experiments in \S\ref{sec:experiments} (Tables~\ref{tab:main}--\ref{tab:structure}) and the ablations in \S\ref{sec:analysis} (Tables~\ref{tab:ablation}, \ref{tab:weight_sensitivity}). The self-evolution specialization finding is supported by Tables~\ref{tab:specialization}, \ref{tab:se_full} and is robust under three random seeds (Appendix~\ref{app:significance}).
\item \textbf{Datasets:} All four evaluation benchmarks are publicly released (ClearFacts~\citep{seo2025vtv}, FEVER~\citep{thorne2018fever}, TruthfulQA~\citep{lin2022truthfulqa}, HaluEval~\citep{li2023halueval}). Statistics in Table~\ref{tab:eval_data}; stratified-sampling protocol in Appendix~\ref{app:data}. Training data is built from ANLI~\citep{nie2020adversarial}; structured annotations are released alongside code.
\item \textbf{Model:} Base models (Qwen2.5-3B-Instruct, Qwen2.5-7B-Instruct) are publicly released under the Apache~2.0 license; fine-tuned checkpoints will be released on publication.
\item \textbf{Training details:} Full hyperparameters for SFT (Table~\ref{tab:sft_hyper}), GRPO (Table~\ref{tab:grpo_hyper}), inference (Table~\ref{tab:inference}), and the four self-evolution rounds (Appendix~\ref{app:se_full}).
\item \textbf{Reward function:} Algorithm~\ref{alg:reward} provides the full process-reward computation; Appendix~\ref{app:reward} gives per-component scoring rubrics.
\item \textbf{Evaluation:} Macro-F1 and accuracy computed against ground-truth labels with greedy decoding (temperature 0); confidence intervals via paired BCa bootstrap with $B{=}10{,}000$ (Appendix~\ref{app:significance}).
\item \textbf{Compute:} 3B-only pipeline reproducible in $\sim$28 GPU-hr on commodity 48\,GB GPUs; full pipeline (4 self-evolution rounds on 7B) in $\sim$130 GPU-hr on a mixed Ada/A100 setup (Tables~\ref{tab:compute}, \ref{tab:carbon}).
\item \textbf{Random seeds:} Seed-to-seed standard deviation reported in Appendix~\ref{app:significance} ($\leq 1.2$\,F1 on auxiliary benchmarks).
\item \textbf{Licenses:} All released artifacts are licensed Apache~2.0 (code) and CC-BY-4.0 (data), consistent with the upstream benchmark licenses.
\item \textbf{Ethics and broader impact:} Discussed in Appendix~\ref{app:ethics}; carbon footprint in Appendix~\ref{app:carbon}.
\end{itemize}

% ---- Reproducibility ----
\section*{Reproducibility Statement}

To ensure reproducibility:
\begin{itemize}[nosep,leftmargin=*]
\item \textbf{Code:} Full training and evaluation code --- including reward functions, GRPO configuration, and data-processing pipelines --- is publicly available at \url{https://github.com/Justin0504/Verifiable_agent}.
\item \textbf{Data:} The structured \seva training data ($4{,}992$ samples), GRPO prompts ($4{,}500$ samples), adversarial probes from the self-evolution loop ($1{,}122$/$2{,}013$/$7{,}787$ for Rounds~2/3/4), and evaluation splits are released alongside the code.
\item \textbf{Models:} We use publicly available base models (Qwen2.5-3B-Instruct and Qwen2.5-7B-Instruct).
\item \textbf{Hyperparameters:} All hyperparameters are listed in Appendix~\ref{app:impl}.
\item \textbf{Compute:} Experiments require $\sim$28 GPU-hours total (Table~\ref{tab:compute}), accessible to academic labs.
\item \textbf{Evaluation:} We report macro F1 and accuracy on standard benchmarks with fixed random seeds. Evaluation uses greedy decoding (temperature 0) for deterministic results.
\end{itemize}

\end{document}